\documentclass[10pt,twocolumn,letterpaper]{article}

\usepackage{wacv}
\usepackage{times}
\usepackage{epsfig}
\usepackage{graphicx}
\usepackage{amsmath}
\usepackage{amssymb}
\usepackage{multirow}
\usepackage{wrapfig,booktabs}
\usepackage{caption}
\usepackage{tabularx}
\usepackage{xcolor}
\usepackage{xr-hyper}
\usepackage[export]{adjustbox}

\def\wacvPaperID{494} 

\wacvfinalcopy 

\ifwacvfinal
\def\assignedStartPage{1} 
\fi

\ifwacvfinal
\usepackage[breaklinks=true,bookmarks=false]{hyperref}
\else
\usepackage[pagebackref=true,breaklinks=true,colorlinks,bookmarks=false]{hyperref}
\fi

\ifwacvfinal
\else
\fi

\title{H2O-Net: Self-Supervised Flood Segmentation via Adversarial Domain Adaptation and Label Refinement}

\begin{document}


\author{\large Peri Akiva\textsuperscript{1} \hspace{0.4cm} Matthew Purri\textsuperscript{1} \hspace{0.4cm} Kristin Dana\textsuperscript{1} \hspace{0.4cm} Beth Tellman\textsuperscript{2} \hspace{0.4cm} Tyler Anderson\textsuperscript{2}\\
\textsuperscript{1}Department of Computer and Electrical Engineering, Rutgers University\\
\textsuperscript{2}Cloud to Street\\
{\tt\small \{peri.akiva, matthew.purri, kristin.dana\}@rutgers.edu \hspace{0.4cm} \{beth, tyler\}@cloudtostreet.info}\\
}

\twocolumn[{%
\renewcommand\twocolumn[1][]{#1}%
\maketitle
\vspace{-2.5em}
\begin{center}
    \centering
    \includegraphics[width=\textwidth]{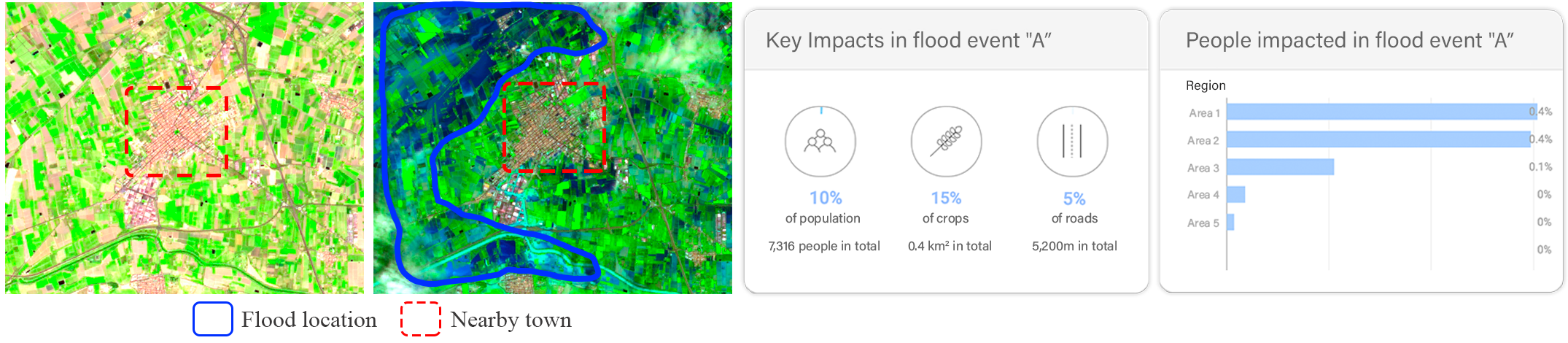}
    \captionof{figure}{Illustration of our method's use-case. Flood segmentation combined with information about per-pixel population density and infrastructure locations (extracted from \cite{tiecke2017mapping}) can help assess damages and response in future events. Numbers in this example are for demonstration purposes and do not reflect real events. }
\end{center}%
}]

\begin{abstract}
    \vspace{-1.05em}
    Accurate flood detection in near real time via high resolution, high latency satellite imagery is essential to prevent loss of lives by providing quick and actionable information. Instruments and sensors useful for flood detection are only available in low resolution, low latency satellites with region re-visit periods of up to 16 days, making flood alerting systems that use such satellites unreliable. 
    %
    %
    This work presents H2O-Network, a self supervised deep learning method to segment floods from satellites and aerial imagery by bridging domain gap between low and high latency satellite and coarse-to-fine label refinement. H2O-Net learns to synthesize signals highly correlative with water presence as a domain adaptation step for semantic segmentation in high resolution satellite imagery. Our work also proposes a self-supervision mechanism, which does not require any hand annotation, used during training to generate high quality ground truth data. We demonstrate that H2O-Net outperforms the state-of-the-art semantic segmentation methods on satellite imagery by 10\% and 12\% pixel accuracy and mIoU respectively for the task of flood segmentation. We emphasize the generalizability of our model by transferring model weights trained on satellite imagery to drone imagery, a highly different sensor and domain.
\end{abstract}
%

\section{Introduction}

Natural disasters cause over 300 billion dollars a year in economic loss, with majority damages done by floods \cite{hallegatte2016unbreakable, guha2015dat}, affecting over 2.3 billion people \cite{wahlstrom2015human}. 
Recent work in poverty dynamics \cite{dang2014remained,baez2015gone} show that natural disasters disproportionately impact poor communities, and push 26 million people into poverty every year \cite{hallegatte2016unbreakable}, with recovery time of up to a decade \cite{dercon2014live}. Difficulties in flood prediction and low response capabilities also make floods the natural disaster with the highest impact rate
%
%
\cite{jonkman2005global,pesaresi2017atlas}, accounting for 57\% of all natural disaster victims \cite{guha2012annual}. The use of satellite for active flood risk assessment and real-time flood response have shown to greatly mitigate flood damages and decreased recovery time \cite{alfieri2018global, oddo_value_2019}. 
%
%

Satellite data comes from two main sources: government and commercial satellites. Government optical satellites tend to be equipped with a large variety of sensors
%
%
(Red, Green, Blue, Coastal aerosol, Red Edge (RE\textsubscript{1}, RE\textsubscript{2}), Near Infra Red (NIR), Narrow NIR, Water vapor, Cirrus, Short Wave Infra Red (SWIR\textsubscript{1}, SWIR\textsubscript{2})) often used in flood detection, but have lower resolution data 
%
%
and have infrequent revisit times over given regions, ranging from 6 to 16 day periods \cite{drusch2012sentinel}. 
%
%
Commercial satellites typically only provide RGB + NIR imagery, but offer higher resolution data with a revisit time of 1 day \cite{murthy2014skysat,anderson2012worldview}. 
%
%
This means that useful data essential for flood detection has low resolution 
%
%
and is only available once every 6 days under optimal conditions,
%
%
making it unusable 
%
%
for real-time flood forecasting and evacuation alarming. 
While commercial satellites 
%
%
have high revisit times essential for real time flood detection, they lack instruments to detect them reliably.
%
%
Note that neither sources provides direct indication of water presence or ground truth data. 
Our method aims to use the best 
%
%
from both worlds through a domain adaptation approach, transferring knowledge from government satellite
data onto commercial satellites data, providing essential instruments for flood segmentation in high resolution data that is available in near real time. 
%
%
To achieve near real time flood segmentation, this work tackles two challenges common in remote sensing and flood detection systems. The first is the lack
of pixel-wise ground truth data and annotation difficulties stemming from complex structures and obstructions such as small streams, clouds, and cloud shadows often occurring in satellite imagery, especially during flood events. Current work \cite{bonafilia2020sen1floods11,cooley_tracking_2017} are able to get coarse masks using classic thresholding methods often used in remote sensing \cite{han2005study}; however, those are often noisy and unreliable for supervision. We solve that by sampling high confidence ground truth points to train a lightweight refiner network on-the-fly and predict the remaining pixels to obtain more reliable source of supervision. The second challenge is detection of floods in high resolution and high 
%
%
temporal frequency satellite imagery. As mentioned before, floods often require near real time response which 
%
%
is only possible in low latency commercial satellites, and those are not equipped with instruments reliable enough to detect water presence.
We address this issue by learning to synthesize SWIR signals 
from low resolution data using an attention driven generative adversarial network, which are concatenated to the input image and fed to a segmentation network supervised by the refined mask.
The primary contributions of this work are as follows:
\begin{itemize}
    \item We propose a method named H2O-Network 
    %
    %
    that learns SWIR signal synthesis in lower resolution data as a domain adaptation mechanism for accurate flood segmentation in high resolution satellite imagery using self-supervision and label refinement. 
    \item  We present a simple run-time refiner that utilizes our adaptive distance map method for coarse-to-fine mask generation. 
    \item We outperform SOTA remote sensing and semantic segmentation methods for flood segmentation.
\end{itemize}


%

\section{Related Work}

\paragraph{Computer Vision and Remote Sensing}

Computer vision and remote sensing commonly align in land cover classification \cite{belgiu_random_2016, van_tricht_synergistic_2018-1}, flood detection \cite{shen_near-real-time_2019-1}, weather prediction \cite{klein2015dynamic}, and precision agriculture \cite{akiva2020finding} using data from government satellites Sentinel-2 and Landsat-8 \cite{sentinelhub} and commercial satellites PlanetScope and WorldView \cite{planet_2020}.
Early and current work use thresholded Normalized Water Index (NDWI) and Modified Normalized Water Index (MNDWI)  \cite{cooley_tracking_2017,xu_modification_2006} for permanent water segmentation, providing quick and somewhat reliable ground truth data for
random forests and support vector machines methods used for land cover classifications \cite{belgiu_random_2016, van_tricht_synergistic_2018-1} and flood detection \cite{shen_near-real-time_2019-1}. 
Recently published remote sensing datasets \cite{schmitt_sen12ms_2019, bonafilia_sen1floods11_2020, alemohammad2019radiant} have encouraged development of deep learning methods which showed improved capabilities in identifying clouds and cloud shadows  \cite{zantedeschi_cumulo:_2019,wieland_multi-sensor_2019}, water \cite{isikdogan_seeing_2019, isikdogan_surface_2017}, inundation \cite{bonafilia2020sen1floods11}, and crop conditions \cite{alcantara2012mapping,akiva2020finding} more accurately than threshold based algorithms when labeled data is available. 
%
%

%
\vspace{-1.5em}
\paragraph{Semantic Segmentation}
Semantic segmentation is the task of assigning class labels to specific pixels in input images. This has been a fundamental computer vision task, with fully connected networks (FCN) \cite{long2015fully} serving as a foundation for semantic segmentation methods in recent years. The encoder-decoder architecture learns optimal interpolation of dense features to recover original resolution and capture scene details. U-Net \cite{ronneberger2015u} is one notable recent method which uses skip connections for better feature propagation similar to \cite{he2016deep}. PSP-Net, DeepLab, and FPN \cite{lin2017feature,chen2017rethinking,zhao2017pyramid} use spatial pyramid pooling to obtain multi scale contextual features useful for segmentation. The importance of contextual features have been further illustrated in \cite{yu2015multi, yu2017dilated}, which expand receptive fields using dilated convolutions, capturing more contextual information without increasing the number of parameters. 
Our approach uses U-Net for the segmentation network, but can practically integrate with any other segmentation methods. 

Interactive segmentation \cite{grady2006random,rother2004grabcut,gulshan2010geodesic,hariharan2011semantic,bai2014error,li2018interactive,xu2016deep,maninis2018deep,jang2019interactive} considers a set of inputs, generally points or scribbles, provided by a user as guidance for semantic segmentation. Classical methods \cite{grady2006random,gulshan2010geodesic,hariharan2011semantic,bai2014error,rother2004grabcut} set this task as an optimization problem, using heuristics and greedy approaches without considering global and local semantic information, while deep learning methods \cite{xu2016deep,liew2017regional,li2018interactive,maninis2018deep, jang2019interactive} utilize global semantic information with respect to the user input. Early work \cite{xu2016deep} stacked input images with distance maps generated from user inputs as network input, essentially generating dense features from sparse input, greatly improving segmentation performance compared to classical methods. Subsequent works commonly follow this approach with added modules such as patch-wise prediction for refinement \cite{liew2017regional}, multi-mask prediction for every object \cite{li2018interactive}, and a seed generator for automatic user input \cite{song2018seednet}. While our work employs elements often seen in interactive segmentation, it removes the interactive part by automatically sampling high confidence points as user input. While this removes the need for user input, it creates potential for class-wise imbalance sampling in cases when an image is fully or almost fully covered with one class. We address this issue by using our adaptive distance map generation that considers the class-wise samples density, allowing more stable predictions. 
\vspace{-1.5em}
\paragraph{Domain Adaptation}
Domain adaptation seeks to transfer knowledge from the labeled source domain to an unlabeled target domain. Traditional domain adaption work has focused on classification and detection tasks \cite{saito2018maximum,tzeng2014deep,ganin2016domain,long2015learning,long2017deep,tzeng2017adversarial}, with more recent work showing advances in semantic segmentation \cite{chen2019domain, pan2020unsupervised,park2019preserving,vu2019advent}. Generally, domain adaptation for semantic segmentation is more complex than for classification and detection since its output often involves highly structured and complex semantic information. 
Most domain adaptation models are comprised of two networks: one for feature adaptation learning, and another for task learning. ADVENT \cite{vu2019advent} improves semantic segmentation through entropy map adaptation to push decision boundaries toward low density regions in target domain. Wu et al. \cite{wu2018dcan} learns to reduce domain shift at both pixel and feature levels using channel-wise feature alignment in image generator and segmentation networks. A similar approach is used in \cite{hoffman2018cycada}, with alignments occurring in both image space and latent space, which are learnt through cycle consistency loss \cite{zhu2017unpaired}. Here, our method learns to generate SWIR signals from our source domain, low  resolution satellite data, and transfer that knowledge to our target domain, high resolution satellite data, to improve semantic segmentation. According to our knowledge, we are the first to achieve this task using only RGB inputs, and first to utilize this method as a domain adaptation step.
\vspace{-1.5em}
\paragraph{Generative Models}
Generative models in the context of domain adaptation has gained traction in recent years. Initial approaches use StyleGAN and Pix2Pix \cite{Isola_2017_CVPR,karras2019style} methods as a data augmentation step, learning domain-invariant features by modifying textures and appearances \cite{hoffman2018cycada,wu2018dcan}. More recent works use image-to-image translation techniques to transfer features available in one domain to another, providing additional data applicable to target domain data \cite{wu2018dcan,chang2019all,ying2019x2ct,pan2020unsupervised}. StandardGAN \cite{tasar2020standardgan} builds on StyleGAN to standardize inputs by enforcing similar styles in satellite data to reduce domain gap.  DUNIT \cite{bhattacharjee2020dunit} combines style and content from source and target images by first generating style features from the style image, and concatenates them to features from content domain as input to a third generator.
%
%
The closest method to our work is S2A \cite{rout2020s2a}, which synthesizes SWIR signals using Wasserstein GAN \cite{arjovsky2017wasserstein} with enlarged receptive fields using high dilation and spatial attention modules. A major difference between our work and S2A is that S2A uses NIR signals during synthesis training and inference. NIR signals greatly correlate with SWIR\textsubscript{2} signals ($r^2(NIR,SWIR_2) \approx 0.82$) \cite{du2016water}, practically diminishing the role of visual queues in the data.
%
%
On the contrary, our approach only uses RGB input for SWIR synthesis, making the task more complex due to a large domain gap, but also more generalizable to other domains that use aerial or satellite imagery and don't contain NIR data. 
Additionally, \cite{rout2020s2a} aims to solely synthesize SWIR\textsubscript{2} signals without inferring on the data, mainly framed as an image-to-image translation algorithm. The paper provides suggestions to applications such as wetland delineation, but incorrectly evaluates its performance for the task. Their results compare ground truth and predicted wetlands obtained by thresholding ground truth and predicted MNDWI, which is composed of synthesized SWIR and green bands. Since the green band is constant in both ground truth and prediction, this metric is redundant and reduces to simply the difference between ground truth and predicted SWIR. Since generating ground truth via thresholding is often noisy, pixel-wise labels become susceptible to common false positives, making it unreliable data to evaluate against. In this work, we define the task of synthesizing SWIR signals as a domain adaption step for flood segmentation in high resolution satellite imagery. We manually annotate a test dataset guided by SWIR signals to empirically evaluate our results.
To our knowledge, this work is a first attempt at learning SWIR signals from visual cues and transferring it to high resolution domain for semantic segmentation. 

\begin{figure*}[t!]
    \centering
    \includegraphics[width=0.9\textwidth]{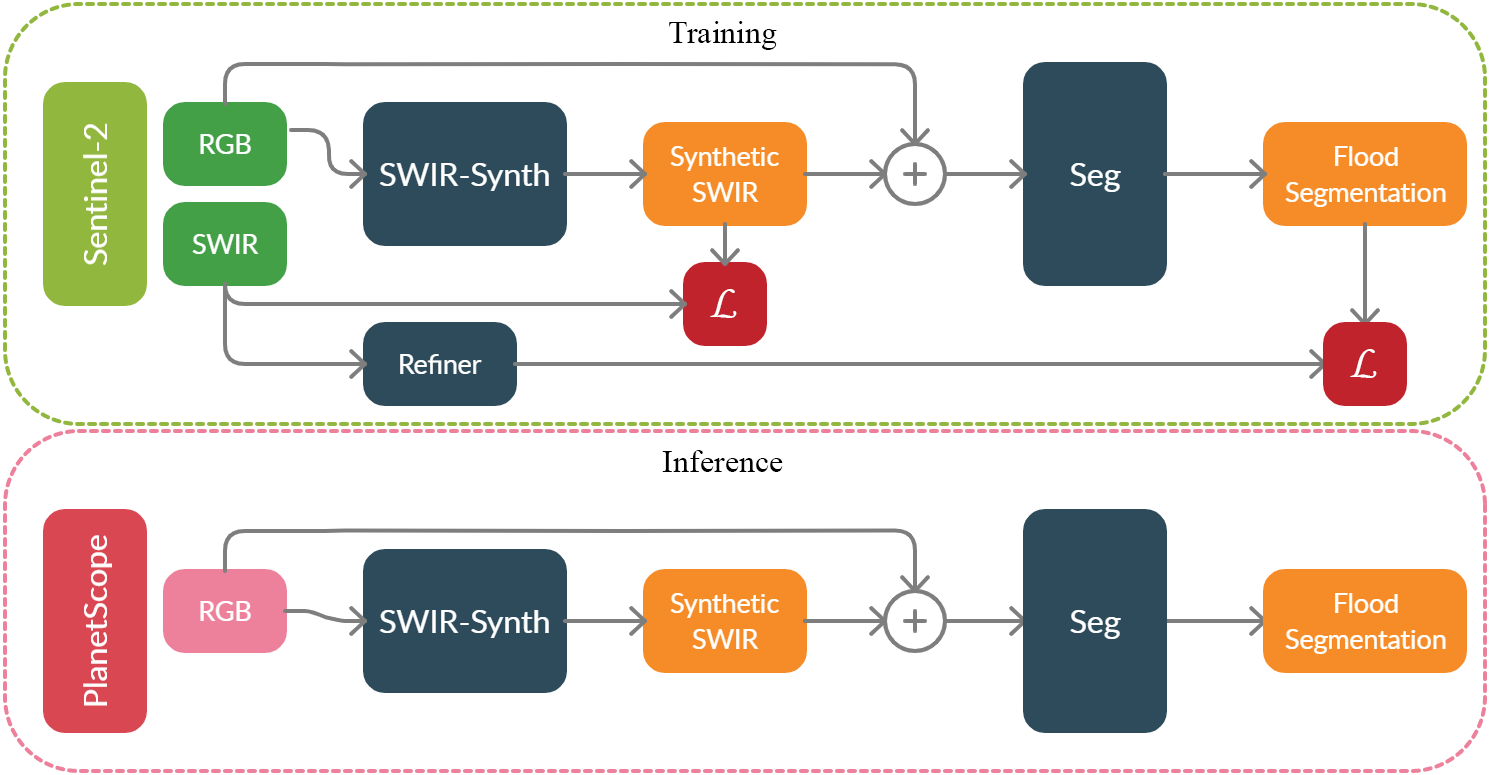}
    \caption{H2O-Net training and inference pipelines. H2O-Net uses Sentinel-2 (low resolution) satellite data to learn synthetic SWIR\textsubscript{2} signals and flood segmentation self-supervised by refined ground truth. It is then evaluated on PlanetScope (high resolution) data. The training pipeline feeds RGB through SWIR-Synth, an adversarial adaptation network, which is then input to the segmentation network, which is self-supervised by refined masks.}
    \vspace{-1.5em}
    \label{fig:my_label}
\end{figure*}

\section{H2O Network}

\subsection{Remote Sensing Overview}
Inundation in medium resolution sensors such as Landsat and Sentinel-2 typically rely on water absorption in the SWIR and NIR bands.
A commonly used water index, MNDWI (the Modified Normalized Water Index, equation \ref{mndwi_eq}) calculates the normalized difference between the green and SWIR bands to emphasize surface water and reduce noise from soil and built up areas \cite{xu_modification_2006}. The MNDWI is a significant improvement over NDWI, which relies on the normalized difference between green and NIR bands \cite{mcfeeters_use_1996}. 
The improvement mainly stems from SWIR\textsubscript{2} unperturbed atmospheric window often occurring in longer wavelength signals as they tend to penetrate the atmosphere more easily.
However, the SWIR\textsubscript{2} band is notably absent from most commercial sensors.
Thresholding, normalized differencing, or more complex combinations of SWIR are used for both Landsat and Sentinel 2 algorithms. In Landsat, SWIR (1560- 1660 nm, Landsat-8 Band 6) and NIR (630-690 nm, Landsat-8 Band 4)  are used with with other bands to detect water \cite{yang_rivwidthcloud:_2019, feyisa_automated_2014, chignell_multi-temporal_2015, donchyts_global_2016, devries_automated_2017}. Algorithms to map surface water for Sentinel-2 similarly rely on water’s absorption in SWIR (1539-1681 nm, Band 11) and NIR (768-796 nm, Band 8) \cite{du2016water}. However, both Landsat and Sentinel-2 still suffer from misclassifications of water and cloud shadows, which both have low reflectance values in SWIR and NIR. The MNDWI is calculated using
\begin{equation}
    \begin{aligned}
    MNDWI = \frac{G - SWIR_{2}}{G + SWIR_{2}} \textit{, }
    \end{aligned}
    \label{mndwi_eq}
\end{equation}
where G represents the green channel.
\subsection{Problem Formulation and Overview}

%
This work aims to learn a mapping between the visual domain $X_{LR} \in \mathbf{R}^{H\times W \times 3}$ of low resolution imagery, and Short Wave Infra Red (SWIR\textsubscript{2}) signals $S \in \mathbf{R}^{H \times W \times 1}$ to improve semantic segmentation prediction, $\Tilde{Y}_{HR} \in \mathbf{R}^{H \times W \times C}$, of flood events in high resolution satellite data, $X_{HR} \in \mathbf{R}^{H\times W \times 3}$, in an unsupervised approach. Since the task of direct segmentation in high resolution imagery, $X_{HR} \Rightarrow \Tilde{Y}_{HR}$, without ground truth is challenging, we bridge the gap by learning signals highly correlative with water presence as a domain adaptation step. During segmentation training, we add a self-supervision method using on-the-fly mask refinement of coarse labels, $Y_{C}$, obtained via classical remote sensing methods, and refined using our lightweight refiner network. During training, the network learns $X_{LR} \Rightarrow \Tilde{S}_{LR} \Rightarrow \Tilde{Y}_{LR}$, which is applied on high resolution data, $X_{HR} \Rightarrow \Tilde{S}_{HR} \Rightarrow \Tilde{Y}_{HR}$. 

\subsection{SWIR-Synth Network}
SWIR-Synth is constructed as an image-to-image translation generative adversarial network where we aim to translate images from the visual domain to the longer wavelengths domain. 
%
Let $x \in \mathbf{R}^{H\times W \times 3}$ be an image from the source domain with an associated SWIR map $s \in \mathbf{R}^{H \times W \times 1}$. The network $G_{SWIR-Synth}$ seeks to synthesize SWIR\textsubscript{2} signals $\Tilde{s} = G_{SWIR-Synth}(x)$ by minimizing the overall loss function
\begin{equation}
    \begin{aligned}
        \mathcal{L}_{GAN}(S,\Tilde{S}) = &\lambda_0\mathcal{L}_{\text{G}}(S,\Tilde{S}) +  \lambda_1\mathcal{L}_{\text{D}}(S,\Tilde{S}) \\ & + \lambda_2\mathcal{L}_{\text{F}}(S,\Tilde{S})\text{,}
    \end{aligned}
\end{equation}
where $\mathcal{L}_{\text{G}}$, $\mathcal{L}_{\text{D}}$, $\mathcal{L}_{\text{F}}$ are the generator, discriminator and feature matching losses, and $\lambda_{\ast}$ represents the weights of losses. As in typical Generative Adversarial Networks (GAN) methods, the generator aims to minimize the objective against an adversarial discriminator that tries to maximize it. 
\vspace{-1em}
\paragraph{Generator Loss}
Generator $G$ learns output $\Tilde{s}$ by minimizing two objective functions: pixel-wise squared error between $s \in S$ and $\Tilde{s} \in \Tilde{S}$, and patch-wise squared error of discriminator output of the generated data. The first objective penalizes inaccurate RGB to SWIR\textsubscript{2} mapping, with the second objective penalize for patches deemed ``fake" by the discriminator. More formally,
\begin{equation}
    \begin{aligned}
        \mathcal{L}_{G}(S,\Tilde{S}) = \mathbf{E}(||s - \Tilde{s}||_{2}) + \mathbf{E}(||1-D(\Tilde{s})||_{2})
    \end{aligned}
\end{equation}
%
%
\paragraph{Adversarial Loss} 
This loss aims to align the features distributions of the two domains by predicting the target labels of the synthesized data. SWIR-Synth is rewarded when it ``fools" the discriminator, resulting in better synthesized output. Here, we utilize a Markovian discriminator (PatchGAN) thoroughly explored in \cite{li2016precomputed}, where the discriminator aims to classify $N\times N$ sub-patches of the generated image, with each patch classified as ``real" or ``fake." The final optimization objective of the discriminator is
\begin{equation}
    \begin{aligned}
        \mathcal{L}_{D}(S,\Tilde{S}) = \mathbf{E}(||D(\Tilde{s})||_{2}) + \mathbf{E}(||1-D(s)||_{2})\textit{.}
    \end{aligned}
\end{equation}
\vspace{-2.2em}
\paragraph{Feature Matching Loss}
Difficult to learn latent patterns often occurring in satellite imagery cause unstable training of the model, which we solve using feature matching loss introduced in \cite{salimans2016improved}. We extract features from discriminator's penultimate layer for both real and synthesized samples, and minimize the squared distance between them. The objective function is
\begin{equation}
    \begin{aligned}
        \mathcal{L}_{F}(S,\Tilde{S}) = \mathbf{E}(||D^{[n-1]}(s) - D^{[n-1]}(\Tilde{s})||_{2}) \textit{,}
    \end{aligned}
\end{equation}
where $n$ represents the number of layers in discriminator $D$, and $D^{[n-1]}(\cdot)$ represents the output of the discriminator's $n-1$ layer. 
\begin{table*}[t!]

\centering
\resizebox{\textwidth}{!}{%
    \begin{tabular}{ccccc}
     \toprule
     Method  & Additional Train Time (sec/iter) & Pixel Accuracy (\%) & mIoU (\%) & FW-IoU (\%) \\
      \midrule
      MNDWI Thresholding ($\leq 0.35$) & \textbf{0} & 84.86 & 76.23 & 79.91  \\
      Refiner (Ours, without Adaptive Distance Maps) & 0.21 & 85.34 & 81.41 & 80.12 \\
      Refiner (Ours, with Adaptive Distance Maps) & 0.36 & \textbf{89.46} & \textbf{84.41} & \textbf{83.04} \\
     \bottomrule\\
\end{tabular}
}
\vspace{-1.5em}
\caption{Mean Intersection over Union (\%), pixel accuracy (\%), and frequency weighted intersection over union (\%) metrics  (higher is better) for refiner network evaluation on manually annotated PlanetScope.}
\label{table:refinermetrics}
\end{table*}

\begin{figure}[t!]
    \centering
    \includegraphics[width=0.5\textwidth]{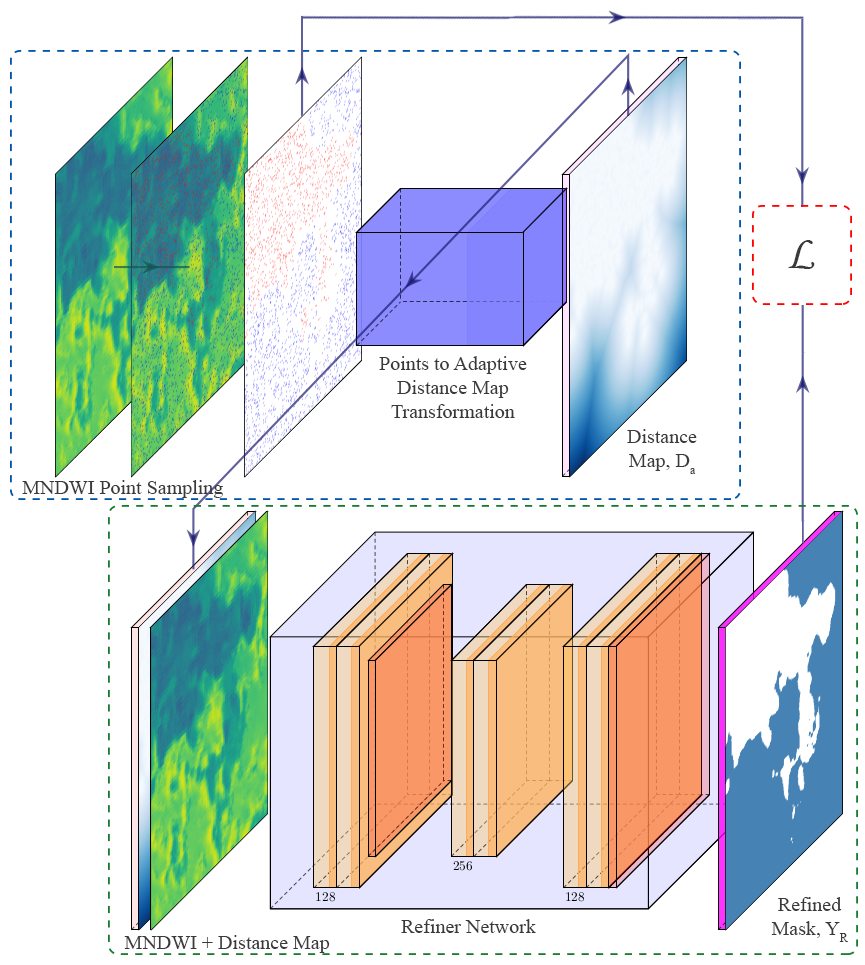}
    \caption{Overview of self-supervision architecture. High confidence points are first sampled from MNDWI and are used to obtain adaptive distance map and as ground truth to the refiner network. Adaptive distance map is concatenated to MNDWI as a 2 channel input to the refiner network. Blue and red dots represent points sampled from water and non-water pixels respectively. Blue and white predicted pixels represent water and non-water pixels. Best viewed in color.}
    \label{fig:refiner}
    \vspace{-1.2em}
\end{figure}
\begin{figure*}[t!]
\setlength\tabcolsep{2pt}
\def\arraystretch{1}
\centering
\begin{tabular}{cccccc}
    \includegraphics[width=0.14\linewidth]{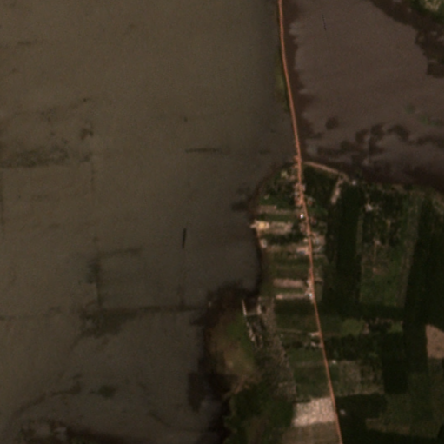}&
    \includegraphics[width=0.14\linewidth]{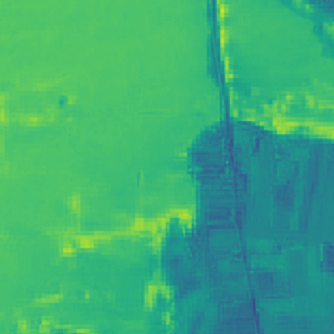}&
    \includegraphics[width=0.14\linewidth]{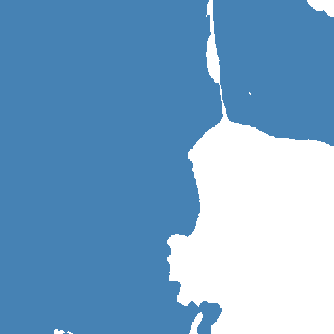} & 
    \includegraphics[width=0.14\linewidth]{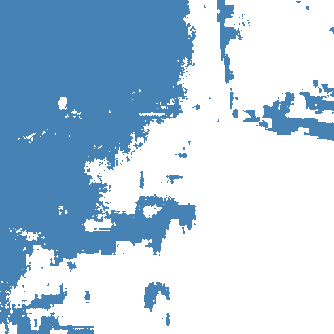} &
    \includegraphics[width=0.14\linewidth]{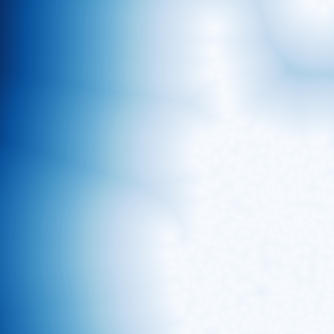} & 
    \includegraphics[width=0.14\linewidth]{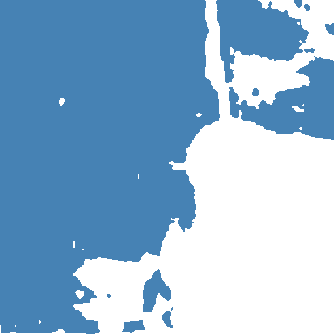} \\
    \includegraphics[width=0.14\linewidth]{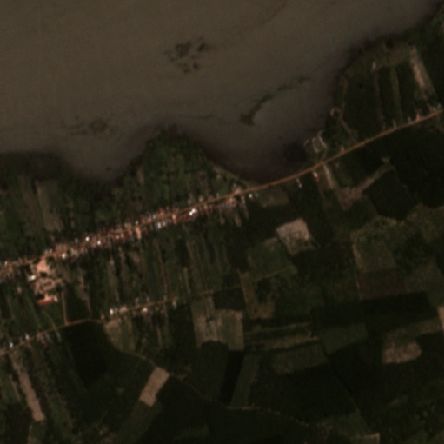}&
    \includegraphics[width=0.14\linewidth]{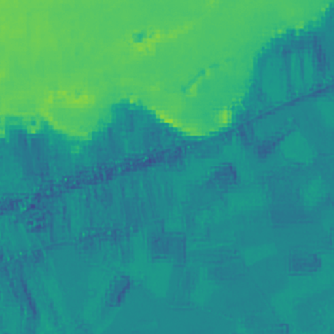} & 
    \includegraphics[width=0.14\linewidth]{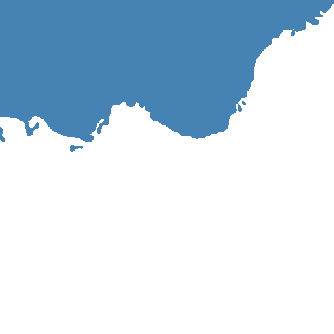} &  
    \includegraphics[width=0.14\linewidth]{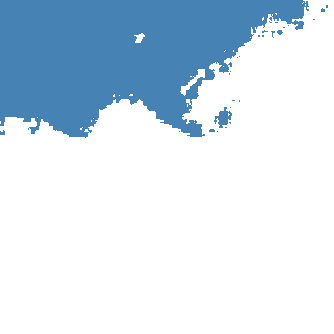} &
    \includegraphics[width=0.14\linewidth]{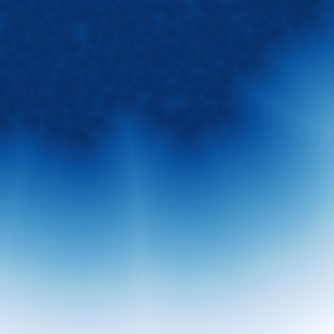} &   
    \includegraphics[width=0.14\linewidth]{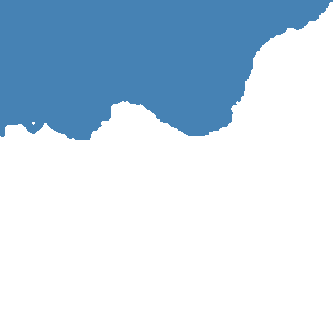} \\
    Image & MNDWI & Ground Truth & MNDWI Threshold & Distance Map & Refined Mask 
\end{tabular}
\vspace{-0.25cm}
\caption{Qualitative results of our refiner on PlanetScope \cite{planet_2020}. Our refiner provides better ground truth for the segmentation network without hand-labeled data. Our refiner samples high confidence points from MNDWI used to generate adaptive distance maps, which are then concatenated to MNDWI to make 2 channel input to the refiner. Blue and white predicted pixels represent water and non-water pixels. Best viewed in color and zoomed.}
\vspace{-0.5cm}
\label{refiner_qualresults}
\end{figure*}

\subsection{Refiner/Self-Supervision}
This work introduces a run time refiner able to generalize over a batch input for the purpose of improving coarse masks obtained via MNDWI thresholding. Let sample $(x \in  \mathbf{R}^{H\times W \times 3} ,m \in \mathbf{R}^{H \times W \times 1})$ be an image-MNDWI pair. We  define high confidence water and non-water pixels using high and low thresholds, $\phi_H$ and  $\phi_L$, on $m$ to obtain a set of locations, $(p_{wx}, p_{wy}) \in P_{w}$ and $(p_{\bar{w}x}, p_{\bar{w}y}) \in P_{\bar{w}} $, corresponding to water (positive) and non-water (negative) points. Note that $|P_w| + |P_{\bar{w}}| \leq |m|$, where $|\cdot|$ is the cardinality of that set. A distance map function is then used to transform $P_w$ and $P_{\bar{w}}$ to distance maps $D_{w} \in \mathbf{R}^{H \times W \times 1} $, and $D_{\bar{w}} \in \mathbf{R}^{H \times W \times 1}$, respectively. Distance maps are calculated by taking the minimum Euclidean distance between a positive point and a set of negative points. For example, to obtain distance map $D_{w}$, we calculate pixel value $D_{w}^{i,j}$ at location $(i,j)$ with the set of points $P_w$ using $D(x,y|P_w) = min_{\forall p_{wx},p_{wy}}\sqrt{(x-p_{wx})^2 + (y-p_{wy})^2}$, where $(x,y)$ is any point in the image. This is repeated until distance values are obtained for all pixels in both $D_{w}$ and $D_{\bar{w}}$.

Traditionally, $D_{w}$ and $D_{\bar{w}}$ would be concatenated to the input image, but this proves to degrade results when the number of sampled points is large. Instead, we dynamically select one distance map that represents the class with highest points density. Class-wise points density of objects is typically comparable in common datasets such as COCO, ADE20K and PASCAL \cite{zhou2017scene, lin2014microsoft, Everingham10}, where objects have similar size and scale. In flood events datasets, classes rarely occur in similar magnitudes, often resulting in sparsely sampled points for one class, and densely sampled points for another class. We define the adaptive distance map, $D_{a}$, as the distance map of the densely sampled points, normalized such that higher values represent the positive class (water), and lower values represent the negative class (no-water). We then construct the network input, $I$, by concatenating MNDWI, $m$, and the adaptive distance map, $D_a$, and use it to train our refiner network, a lightweight convolution refining network, supervised by $P_w$ and $P_{\bar{w}}$, with the rest of the pixels ignored. The network minimizes
\begin{equation}
    \mathcal{L}_{R}(P_w,P_{\bar{w}},\Tilde{y})=-\sum_{p_w} log(\Tilde{y}) - \sum_{p_{\bar{w}}} (1-log(\Tilde{y})) \text{.}
\end{equation}
Note that the loss function ignores pixels not in the set $P_w \cup P_{\bar{w}}$.
After the network is trained for $k$ iterations, it predicts on the same image to obtain mask values for the rest, previously ignored pixels, to produce a pseudo-mask ground truth $\Tilde{Y}_{R}$ fed to the segmentation network. The intuition behind this method is to predict the labels of the entire image given a small set of high confidence labeled pixels. 



\subsection{Segmentation Network}
An encoder-decoder architecture takes a concatenation of RGB image $X$ and synthesized SWIR $\Tilde{S}$ as a 4 channel input, and aims to predict pixel-wise labels $\Tilde{Y}$ that minimize the cross entropy loss between $\Tilde{Y}$ and $Y_{R}$. We define the segmentation loss by
\begin{equation}
    \mathcal{L}_{S}(Y_R,\Tilde{Y})=-\sum log(\Tilde{Y}) - \sum (1-log(\Tilde{Y})) \text{.}
\end{equation}
%

In practice, this network has two roles: it fine-tunes SWIR-Synth by penalizing low synthesized SWIR values (water presence) when water labels are not present, and it learns to correlate RGB and SWIR pairs input to water pixels.

\section{Experiments}
\begin{figure*}[t!]
\setlength\tabcolsep{1pt}
\def\arraystretch{0.5}
\centering
\begin{tabular}{cccccccc}
    \includegraphics[width=0.12\linewidth]{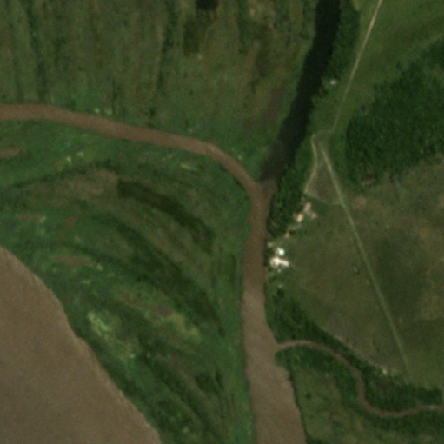} &   \includegraphics[width=0.12\linewidth]{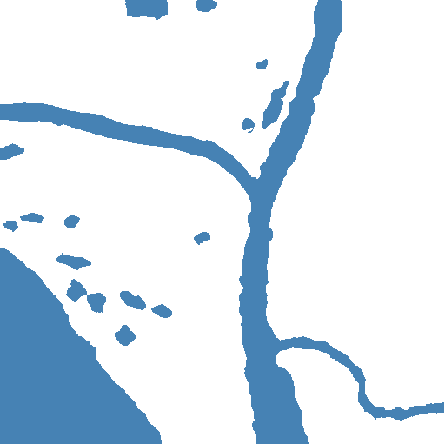} &   \includegraphics[width=0.12\linewidth]{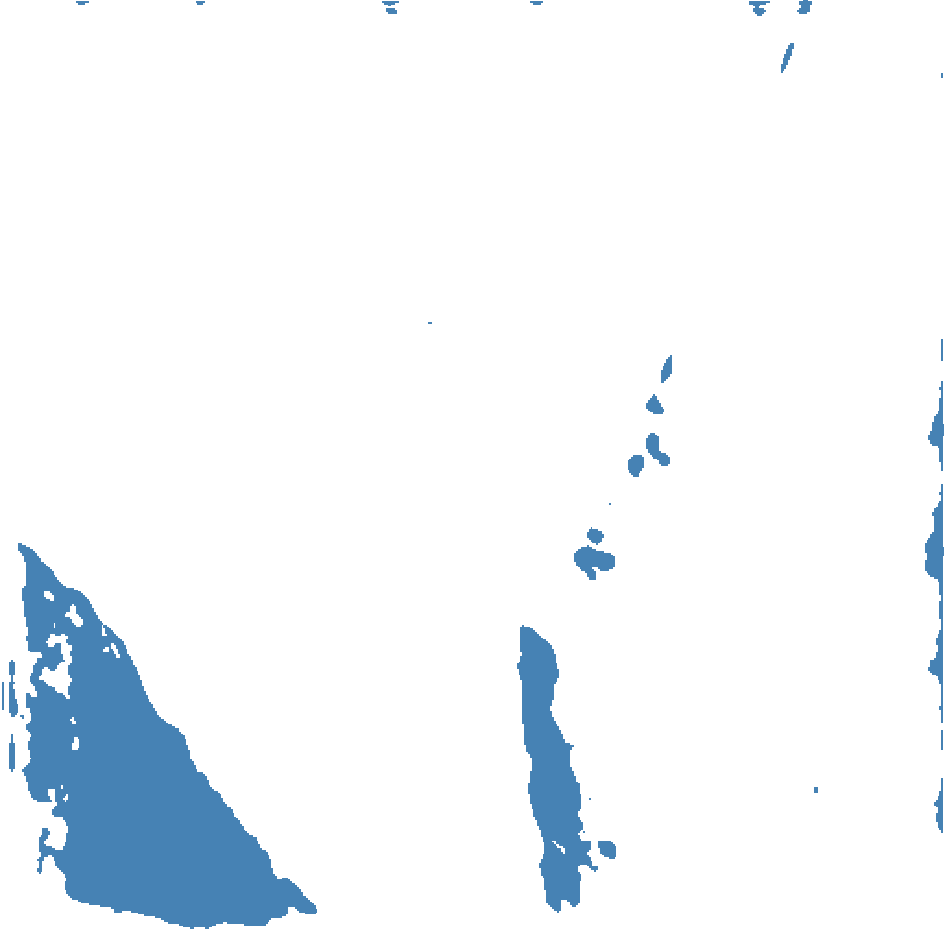} & 
    \includegraphics[width=0.12\linewidth]{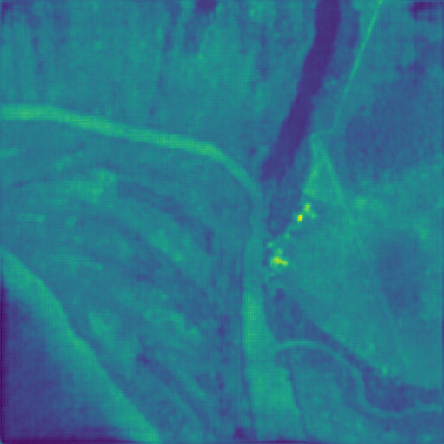} & 
    \includegraphics[width=0.12\linewidth]{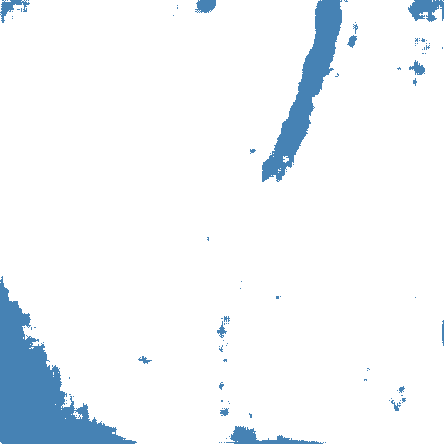} &
    \includegraphics[width=0.12\linewidth]{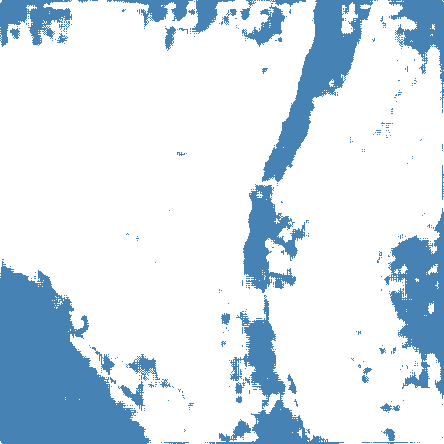} &
    \includegraphics[width=0.12\linewidth]{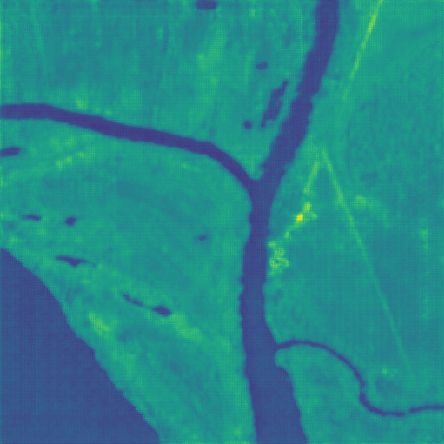} &
    \includegraphics[width=0.12\linewidth]{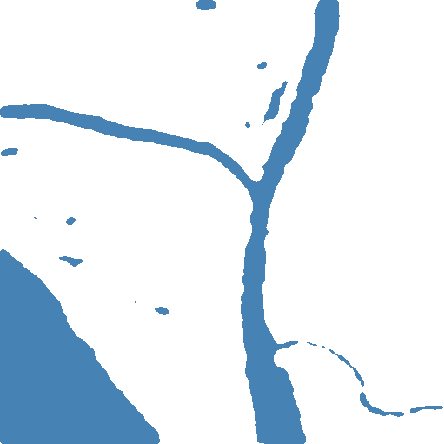}\\
    \includegraphics[width=0.12\linewidth]{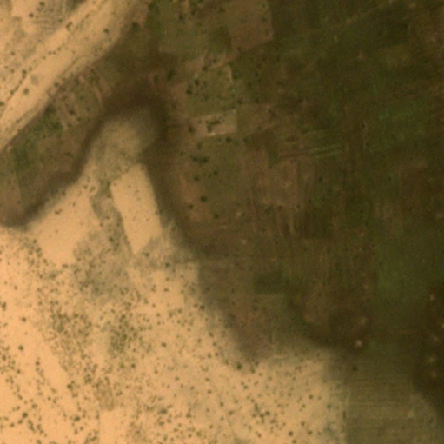} &   \includegraphics[width=0.12\linewidth]{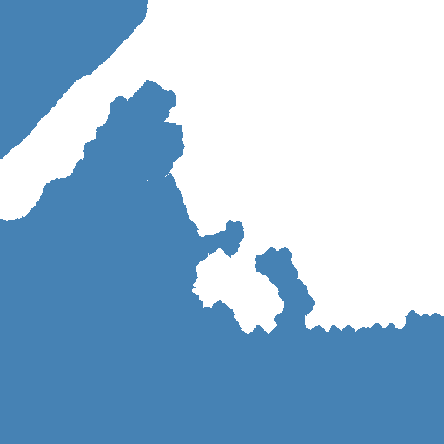} &   \includegraphics[width=0.12\linewidth]{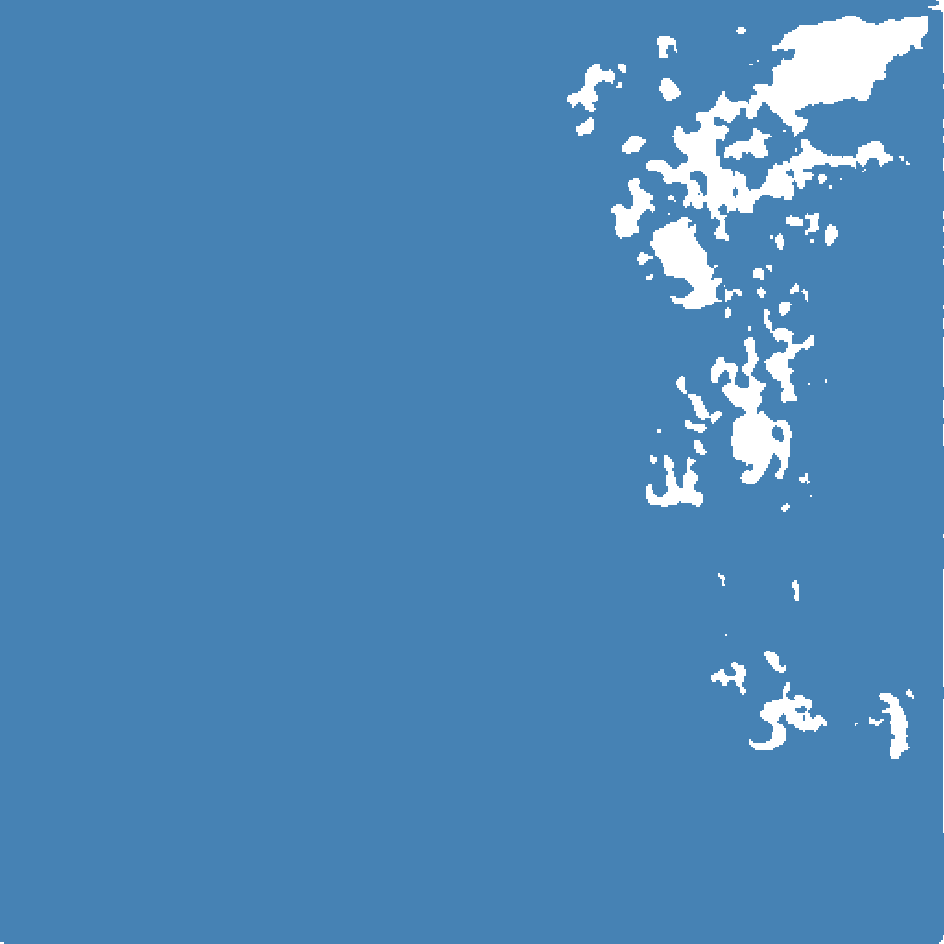} & 
    \includegraphics[width=0.12\linewidth]{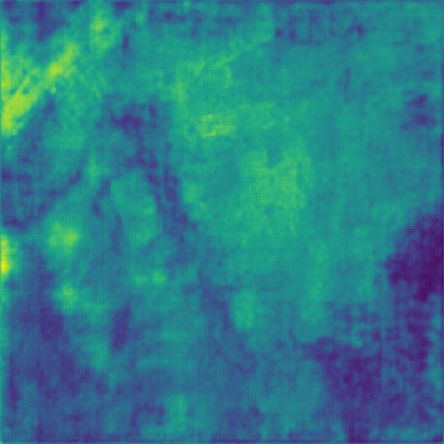} & 
    \includegraphics[width=0.12\linewidth]{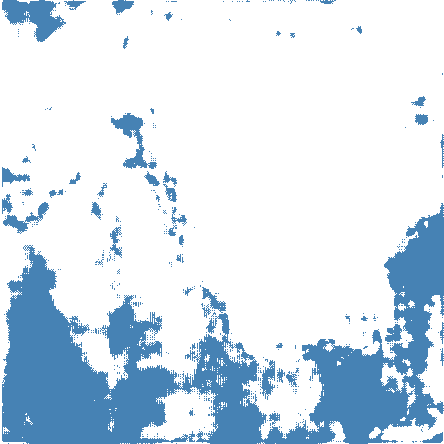} &
    \includegraphics[width=0.12\linewidth]{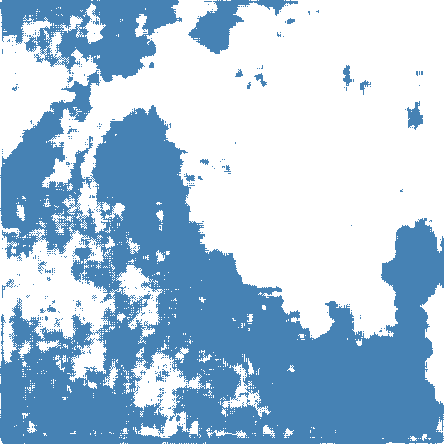} &
    \includegraphics[width=0.12\linewidth]{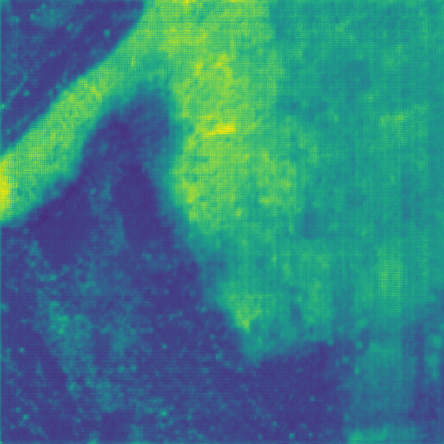} &
    \includegraphics[width=0.12\linewidth]{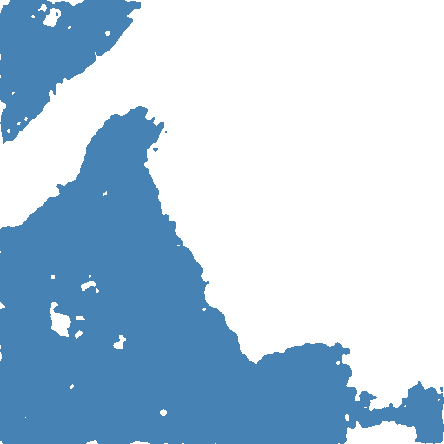}\\
    \includegraphics[width=0.12\linewidth]{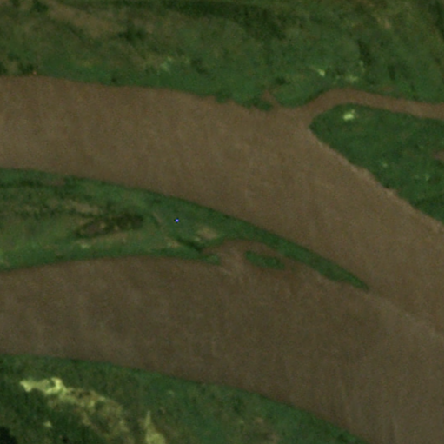} &   \includegraphics[width=0.12\linewidth]{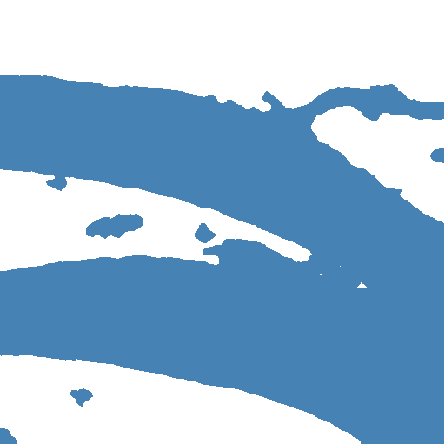} &   \includegraphics[width=0.12\linewidth]{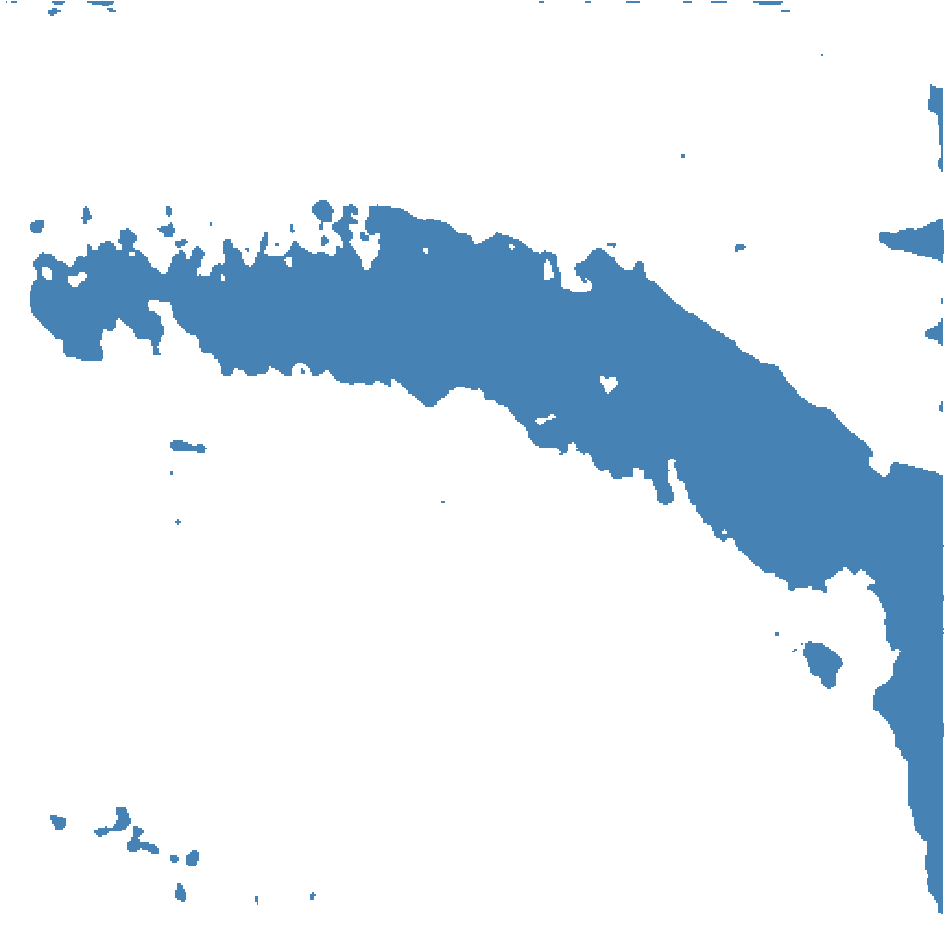} & 
    \includegraphics[width=0.12\linewidth]{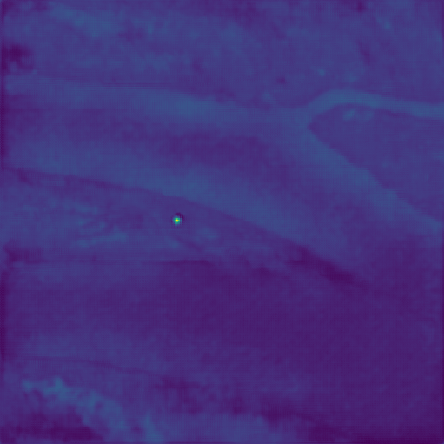} & 
    \includegraphics[width=0.12\linewidth]{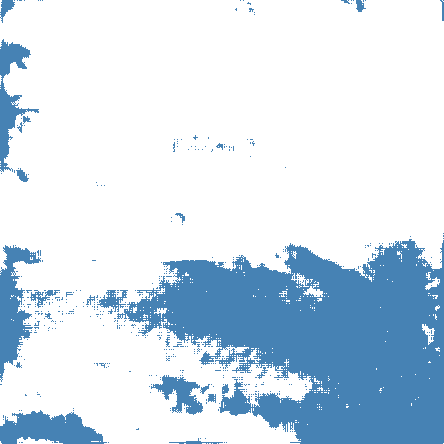} &
    \includegraphics[width=0.12\linewidth]{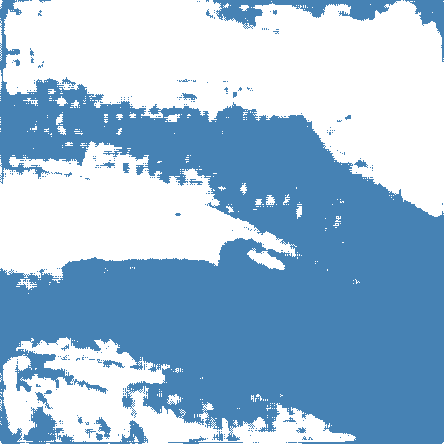} &
    \includegraphics[width=0.12\linewidth]{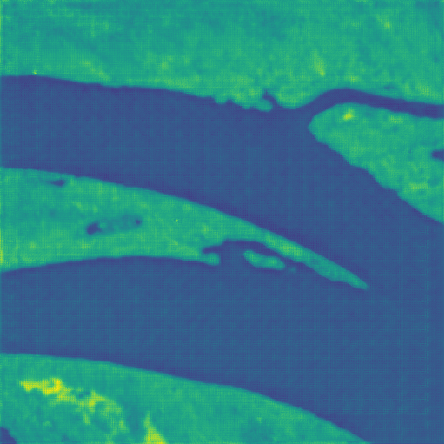} &
    \includegraphics[width=0.12\linewidth]{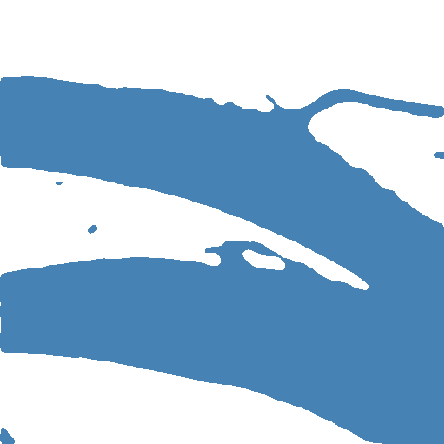}\\
    \includegraphics[width=0.12\linewidth]{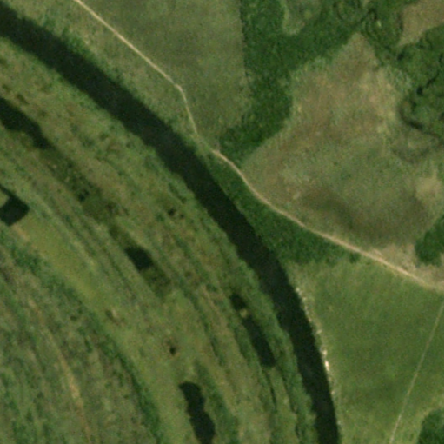} &   \includegraphics[width=0.12\linewidth]{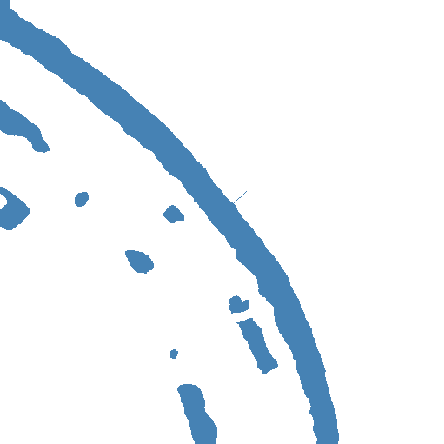} &   \includegraphics[width=0.12\linewidth]{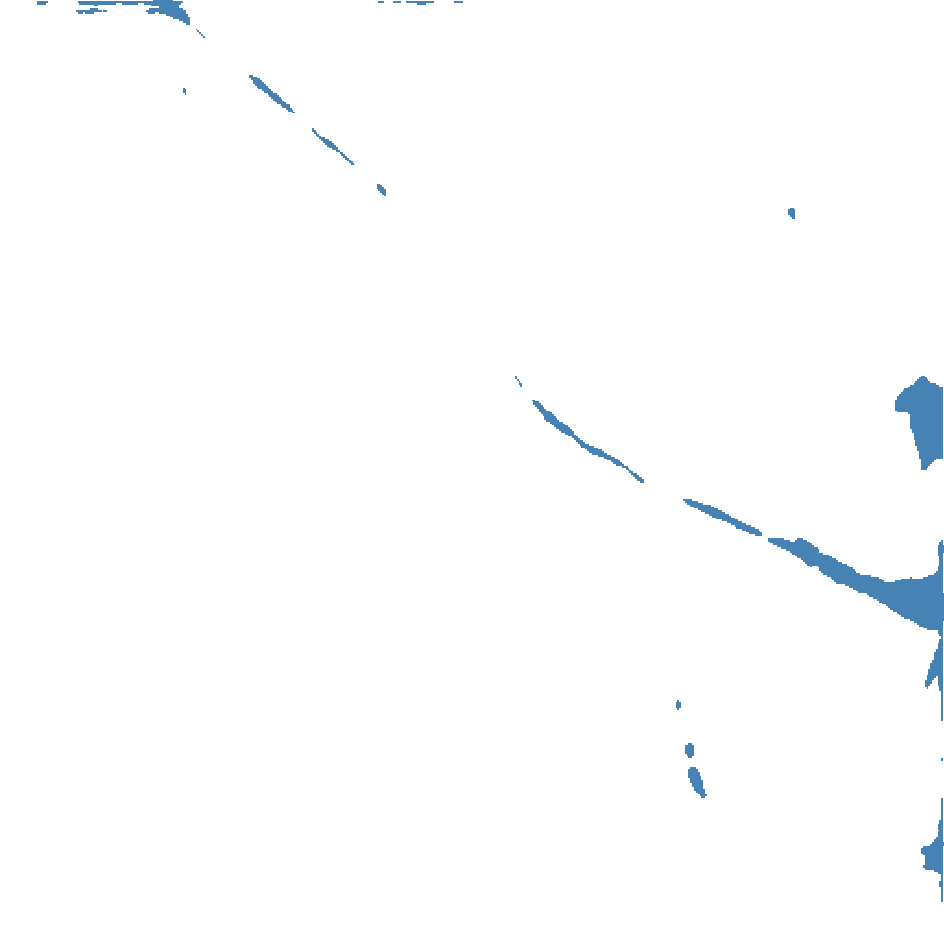} & 
    \includegraphics[width=0.12\linewidth]{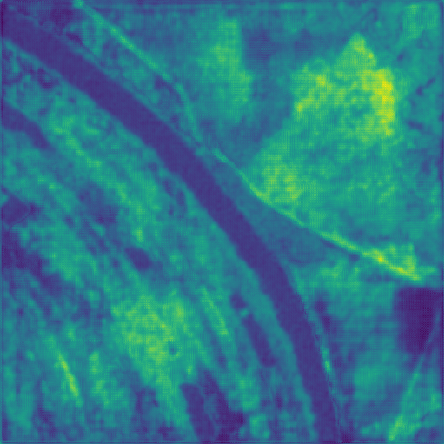} & 
    \includegraphics[width=0.12\linewidth]{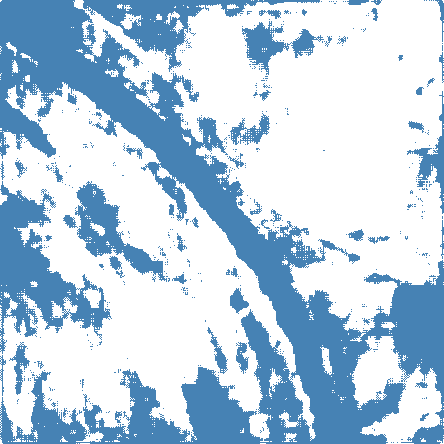} &
    \includegraphics[width=0.12\linewidth]{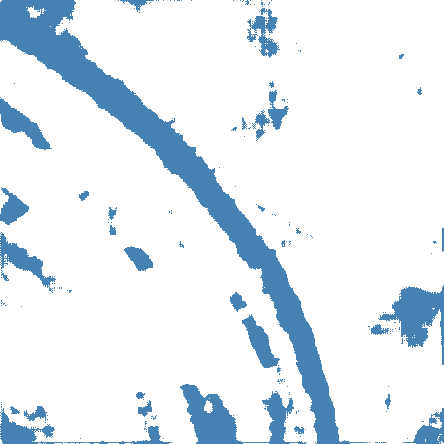} &
    \includegraphics[width=0.12\linewidth]{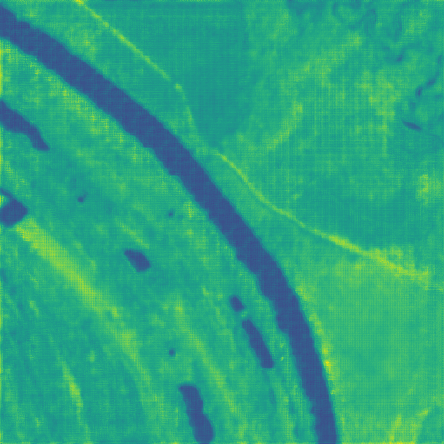} &
    \includegraphics[width=0.12\linewidth]{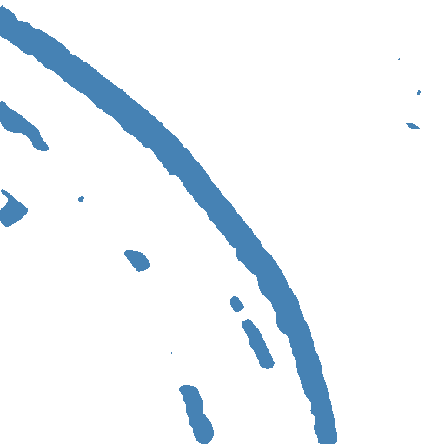}\\
    Input Image & Ground Truth  & \multicolumn{1}{m{2cm}}{\centering U-Net w/ Refiner} & SWIR-Synth & \multicolumn{1}{m{2cm}}{\centering SWIR-Synth Thresh.}  & \multicolumn{1}{m{2cm}}{\centering H2O-Net (RGB)} & \multicolumn{1}{m{2cm}}{\centering SWIR-Synth (RGB+NIR)} & \multicolumn{1}{m{2cm}}{\centering H2O-Net (RGB+NIR)}
\end{tabular}
\vspace{-0.5em}
\caption{Qualitative comparison with SOTA methods on PlanetScope \cite{planet_2020}. H2O-Net results shows that synthesizing SWIR data allows robust segmentation performance in high resolution imagery. It can be seen that adding NIR signals to training and inference improves results in both for segmentation and synthesized SWIR. Blue and white predictions correspond to water and non-water pixels. Best viewed in color and zoomed.}
\vspace{-1.8em}
\label{qualresults}
\end{figure*}
\subsection{Training and Evaluation Setup}
We train all networks from scratch using Adam optimizer \cite{kingma2014adam} with starting learning rates of 2e-4 for generator, 6e-4 for discriminator, 1e-3 for segmentor, and 1e-2 for refiner. We set $\beta$'s for the Adam optimizer to 0.4 and 0.99 and cosine annealing scheduler \cite{loshchilov2016sgdr} for all networks. Training data used is split 90/5/5 with normalization transformation with zero mean and unit variance applied before input. The generator is first trained for 5 epochs before adding the discriminator, which are trained together for an additional 30 epochs. The segmentation network is then added to the network to train jointly for another 60 epochs. For performance evaluation we report Mean Intersection over Union (mIoU), Pixel Accuracy (PA), and Frequency Weighted Intersection over Union (FW-IoU). The metrics are calculated as follows
\begin{equation}
    \begin{aligned}
        &\text{PA} =  \frac{\sum_{i} n_{ii}}{\sum_{i} t_i} \text{ ,}\\
        &\text{mIoU} = \frac{1}{n_{c}} \sum_{i} \frac{n_{ii}}{t_i + \sum_{j} n_{ji} - n_{ii}} \text{ ,}\\
         &\text{FW-IoU} = (\sum_{k} t_k)^{-1} \sum_{i} \frac{t_in_{ii}}{t_i + \sum_jn_{ji}-n_{ii}} \text{ ,}\\
    \end{aligned}
    \label{metrics}
\end{equation}
where $n_{ij}$ is the number of pixels of class $i$ predicted as class $j$, $t_i$ is total number of pixels for class $i$, and $n_c$ is the number of classes.
\subsection{Datasets}
\label{section:datasets}
We train our method and baselines on Sentinel 2 satellite data which is an optical satellite providing RGB, NIR, Narrow NIR, Water vapor, Cirrus, SWIR\textsubscript{1}, and SWIR\textsubscript{2} signals at 10 to 60 meters per pixel. The training dataset is comprised of 3676 images of size 512$\times$512 containing RGB and SWIR\textsubscript{2} data resampled to 10 meters per pixel.
%
%
For evaluation we use PlanetScope \cite{planet_2020} and Drone Deploy \cite{dronedeploy} datasets. PlanetScope is a high resolution optical satellite providing RGB and NIR signals at 3 meter per pixel. In order to evaluate both our refiner network, and H2O-Net on the same data, we obtain overlapping crops in both Sentinel-2 and PlanetScope to obtain SWIR\textsubscript{2} signals measured by Sentinel-2, resampled and stacked with PlanetScope data. Note that this overlap is not a common occurrence and only provided with 246 evaluation images, which were then manually annotated. 
%
%
%
Drone Deploy \cite{dronedeploy} is a high altitude fully annotated aerial imagery dataset with 55 orthomosaic images with 10 centimeters per pixel resolution. We generate 775, 512$\times$512 crops that contain at least 10\% water pixels, which are split to 50/20/30 for fine-tuning, validation and testing sets. 
%
%
\subsection{Baselines}
\label{section:baselines}
This work explores methods relevant to remote sensing and semantic segmentation. Since there is no other work that segments water from satellite imagery without supervision, we compare the closest possible methods. From the remote sensing domain, we use \cite{cooley2017tracking} as a baseline since they use the same evaluation dataset, PlanetScope. It utilizes NDWI thresholding, using NIR signals available in PlanetScope, which are not used in the deep learning baselines. Although the additional information gives NDWI and MNDWI thresholding methods an advantage, we are interested in what can be achieved without this information. Of the semantic segmentation methods, we show U-Net \cite{ronneberger2015u} and DeepLabv3 \cite{chen2017rethinking}. Training those methods follow similar approach to H2O-Net, where we train the networks on low resolution imagery, and predict on high resolution imagery. We include performances of those methods supervised by thresholded MNDWI (coarse labels) and refined labels. All methods were trained from scratch to ensure fair comparison. For methods reporting on DroneDeploy dataset (more details in section \ref{section:ablation}), we fine-tune the models using the given ground truth data before evaluation. Note that all methods were trained from scratch to ensure fair comparison.
\begin{table*}[t]
\centering
\resizebox{0.85\textwidth}{!}{%
\begin{tabular}{@{\hspace{2em}}ccccc@{\hspace{2em}}}
\toprule
Dataset & \multicolumn{4}{c}{PlanetScope \cite{planet_2020}}  \\ \midrule
Method & Bands Used & Pixel Accuracy (\%) & mIoU (\%) & FW-IoU (\%) \\ \midrule
NDWI Thresholding ($\geq -0.1$) \cite{cooley2017tracking,mcfeeters_use_1996} & RGB+NIR& 72.35 & 63.26 & 61.53  \\
SWIR-Synth Thresholding ($\geq 0.35$) & RGB+NIR & 85.47 & 71.93 & 74.49  \\
H2O-Net (Ours) & RGB+NIR & \textbf{87.23} & \textbf{74.52} & \textbf{76.88} \\
\midrule
DeepLab v3 \cite{chen2017rethinking} & RGB & 57.03 & 31.46 & 29.61  \\
DeepLab v3 (w/ refiner)  & RGB & 59.22 & 33.74 & 39.89 \\
U-Net \cite{ronneberger2015u} & RGB & 64.01 & 42.63 & 49.49  \\
U-Net (w/ refiner)  & RGB & 66.81 & 48.18 & 50.98  \\
SWIR-Synth Thresholding ($\geq 0.35$) & RGB & 69.23 & 53.31 & 53.97  \\
H2O-Net (Ours) & RGB &  \textbf{76.19} &  \textbf{60.92} &  \textbf{62.42} \\
\bottomrule
\end{tabular}%
}
\vspace{-0.2 em}
\caption{Pixel accuracy (\%), Mean Intersection over Union (\%), and Frequency-Weighted Intersection over Union (\%) accuracy (higher is better) on high resolution satellite imagery PlanetScope dataset. Reported performance is with respect to water segmentation. It is evident from the table that when we train our method with both RGB and NIR signals, we significantly outperform classical methods that also use NIR signals.}
\vspace{-1.2em}
\label{tab:results}
\end{table*}
\section{Results}
Figure \ref{qualresults} shows the qualitative results for the task of semantic segmentation. Our method, H2O-Net, shows the most consistent results out of the compared baselines. Table \ref{tab:results} shows the quantitative performance of our method compared to SOTA methods in semantic segmentation and remote sensing. Figure \ref{qualresults} shows that H2O-Net is able to finely segment small streams and muddy flood waters compared to isolatedly trained segmentation network. It can be seen that narrowing the domain gap through synthesized SWIR\textsubscript{2} can significantly increase performance. For the refiner, figure \ref{fig:refiner} and table \ref{table:refinermetrics} show superior performance both qualitatively and quantitatively for obtaining self-supervised ground truth data. It is evident from the results that training on coarse labels alone is not enough to provide reliable segmentation prediction in high resolution imagery, and adding the refined masks for supervision improves performance overall. While thresholding methods used in remote sensing performed relatively well compared to deep learning methods, it is important to emphasize that explicit NIR and SWIR\textsubscript{2} reflectance data of the scene was available during evaluation, which is not available for deep learning methods. For that, we also include metrics of our method trained and evaluated with NIR signals. It can be seen that such signals provides significant boost in performance both qualitatively and quantitatively. Since DroneDeploy \cite{dronedeploy} does not provide such data, those methods are also not reported for that dataset. Additional qualitative results and implementation details for H2O-Net, SWIR-Synth, refiner, and baseline networks are available in the supplementary. 

\subsection{Domain Transferability Study}
\label{section:ablation}
We explore the effect of SWIR-Synth and H2O-Net on aerial imagery. The domain gap between low resolution satellite imagery and high resolution aerial imagery (10 meters/pixel compared to 10 centimeters/pixel) is even larger than the domain gap between low and high resolution satellite imagery (10 meters/pixel compared to 3 meters/pixel), making it a more challenging task. Additionally, this dataset does not have flood related imagery, as it was collected over parks and residential areas featuring pools, fountains, and lakes, which inherently have different structures than flood events imagery. Another challenge is in data representation domain gap, as satellite data measures reflectance values for RGB and SWIR\textsubscript{2} signals, while Drone Deploy provides digital values (0-255). Figure 2 in supplementary material and table \ref{tab:dd_results} show the qualitative and quantitative results on DroneDeploy \cite{dronedeploy} dataset after fine-tuning the segmentation network. The results show that despite such domain gap, our method still improves accuracy for water segmentation. In supplementary material we show challenging examples occurring in urban areas. Asphalt and roof-tops often have similar SWIR\textsubscript{2} responses to water, which is inherently also learned to be generated by the SWIR-Synth output.
%

\begin{table}[t]
\centering
\resizebox{\columnwidth}{!}{%
\begin{tabular}{@{}cccc@{}}
\toprule
Dataset &  \multicolumn{3}{c}{Drone Deploy \cite{dronedeploy}} \\ \midrule
Method $\backslash$ Metric  & PA (\%) & mIoU (\%) & FW-IoU (\%) \\ \midrule
DeepLab v3 \cite{chen2017rethinking} & 88.14 & 56.62 & 77.92 \\
U-Net \cite{ronneberger2015u} & 91.11 & 68.99 & 83.64 \\
SWIR-Synth Thresholding ($\geq 0.35$) & 59.31 & 39.24 & 48.37 \\
H2O-Net (Ours) & \textbf{92.71} & \textbf{75.25} & \textbf{86.65} \\
\bottomrule
\end{tabular}%
}
\vspace{-0.5em}
\caption{Domain Transferability Study. Pixel accuracy (\%), Mean Intersection over Union (\%), and Frequency-Weighted Intersection over Union (\%) accuracy (higher is better) on DroneDeploy, a high resolution aerial imagery  dataset. Reported performance is with respect to water segmentation.}
\vspace{-1.8em}
\label{tab:dd_results}
\end{table}

\subsection{Conclusion}

This work presents a novel approach to address difficulties in flood response time and support, allowing detection of floods in high resolution, high temporal frequency satellites without explicit annotation efforts. We propose the H2O-Network which employs SWIR\textsubscript{2} synthesis as a domain adaptation step for semantic segmentation using a self-supervised approach. We hope that this work will provide additional route to save lives and property in communities susceptible to such natural disasters. Our approach also shows that detecting and segmenting objects that have reflectance properties may benefit from such domain adaptation step seeking to gain additional features in the target domain. Important to note that ideally, such application in practice would make use of NIR signals which is often available in commercial satellites, but we choose to omit such data to demonstrate our approach on datasets with a larger domain gap.

\newcolumntype{Y}{>{\centering\arraybackslash}X}

\newpage
{\small
\bibliographystyle{ieee_fullname}
\bibliography{egbib}
}

\end{document}


\title{H2O-Net: Self-Supervised Flood Segmentation via Adversarial Domain Adaptation and Label Refinement Supplementary Material}


\maketitle

\vspace{-1cm}
\section{Additional Qualitative Results}
\begin{figure*}[t!]
\setlength\tabcolsep{1pt}
\def\arraystretch{0.5}
\centering
\begin{tabular}{cccccccc}
    \includegraphics[width=0.12\linewidth]{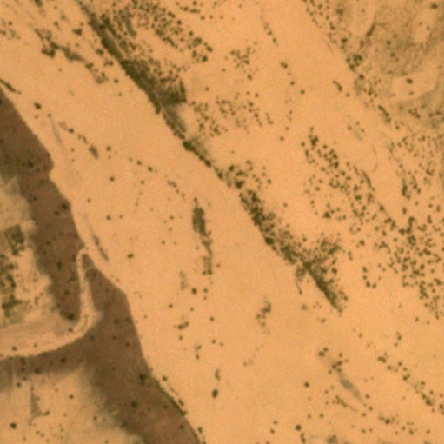} &   \includegraphics[width=0.12\linewidth]{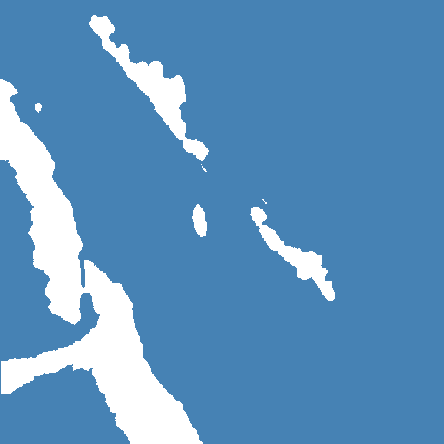} &   \includegraphics[width=0.12\linewidth]{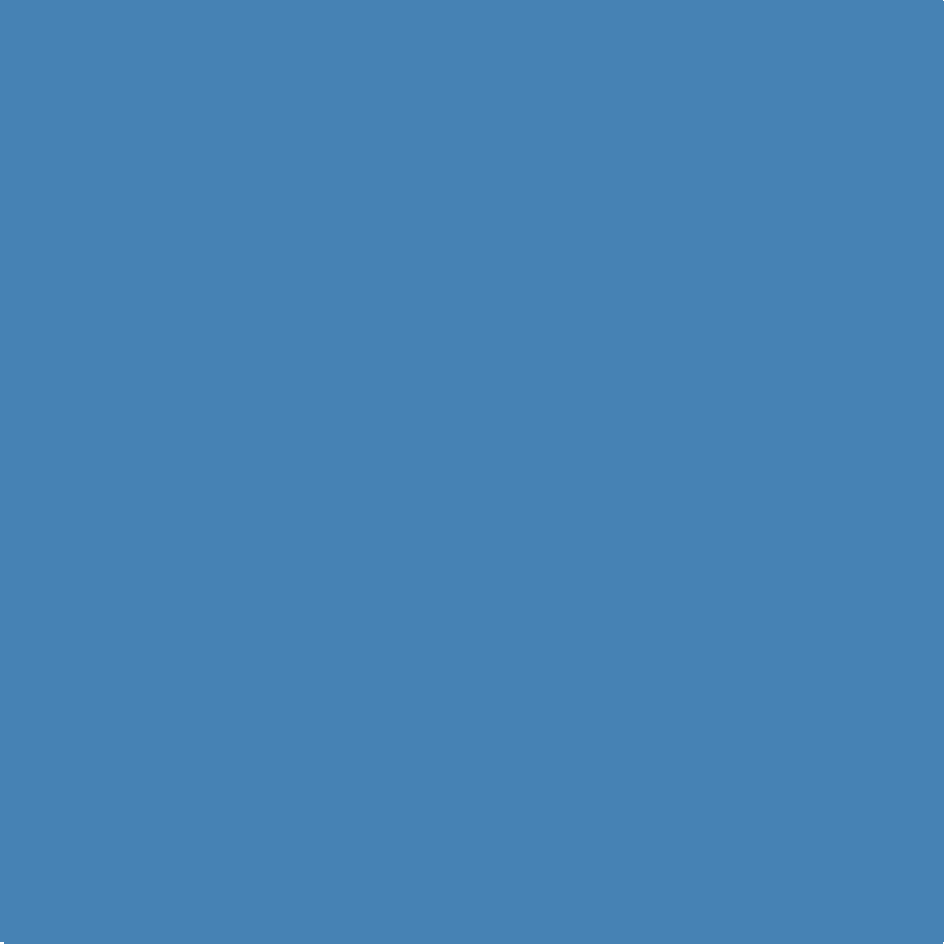} & 
    \includegraphics[width=0.12\linewidth]{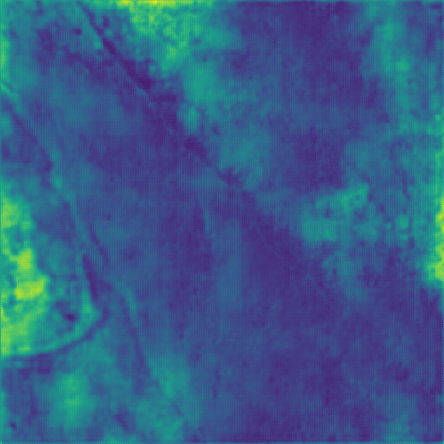} & 
    \includegraphics[width=0.12\linewidth]{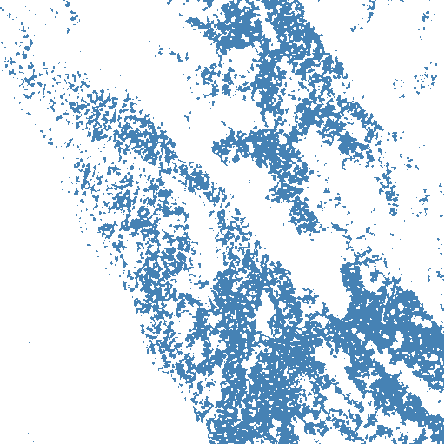} &
    \includegraphics[width=0.12\linewidth]{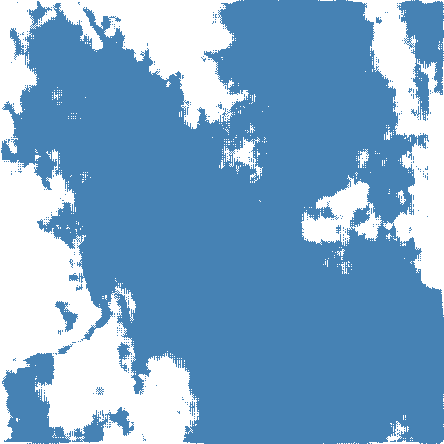} &
    \includegraphics[width=0.12\linewidth]{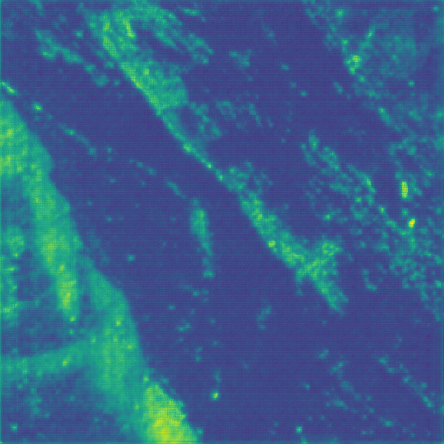} &
    \includegraphics[width=0.12\linewidth]{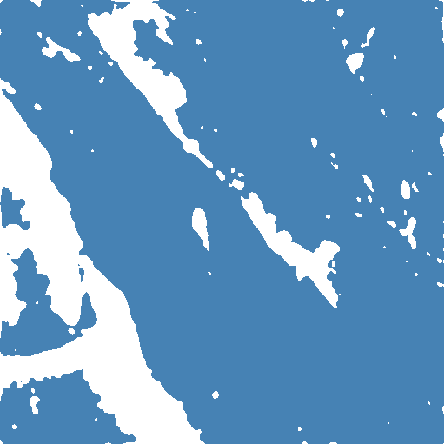}\\
    \includegraphics[width=0.12\linewidth]{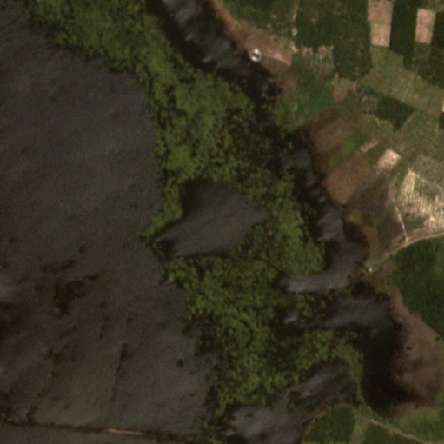} &   \includegraphics[width=0.12\linewidth]{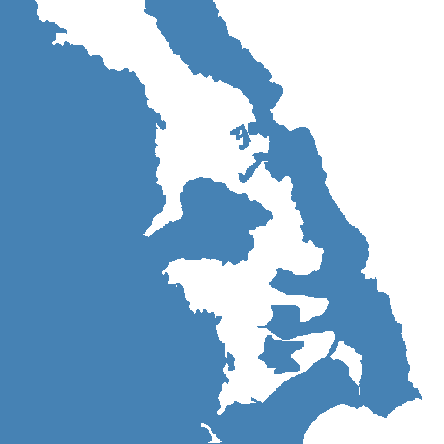} &   \includegraphics[width=0.12\linewidth]{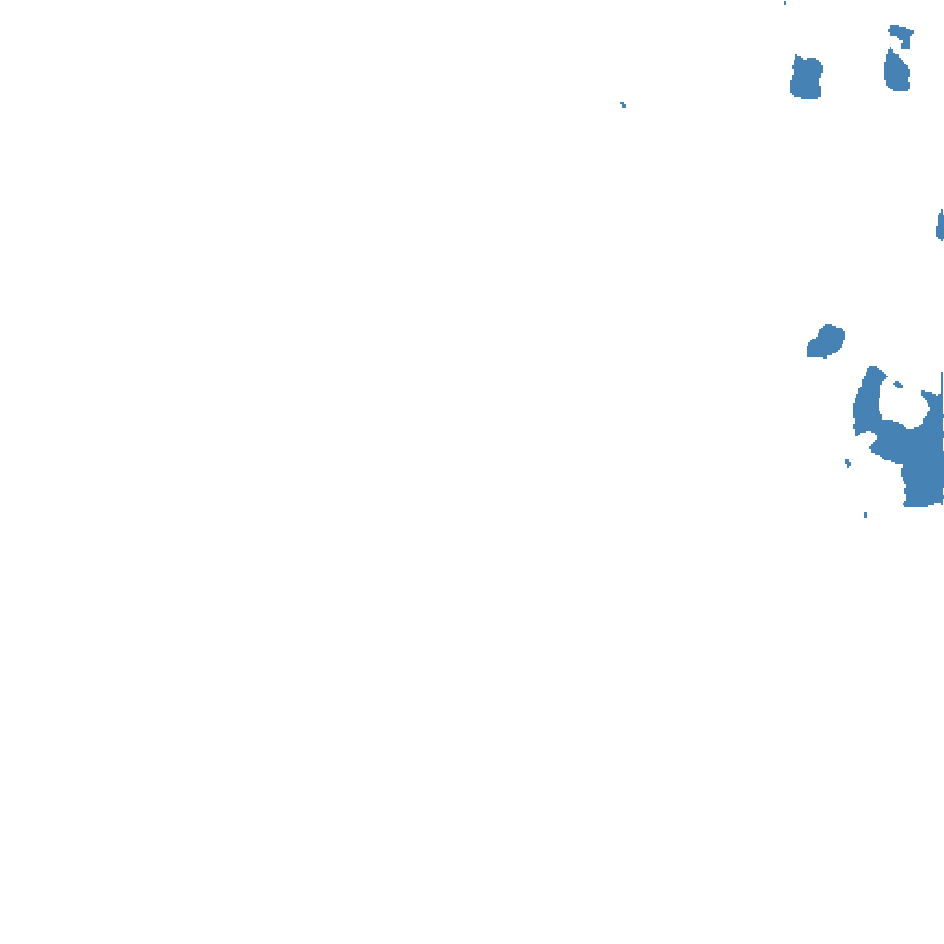} & 
    \includegraphics[width=0.12\linewidth]{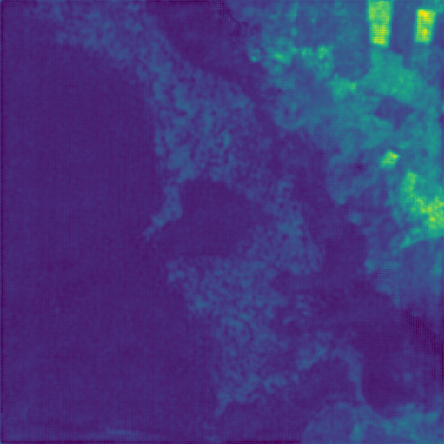} & 
    \includegraphics[width=0.12\linewidth]{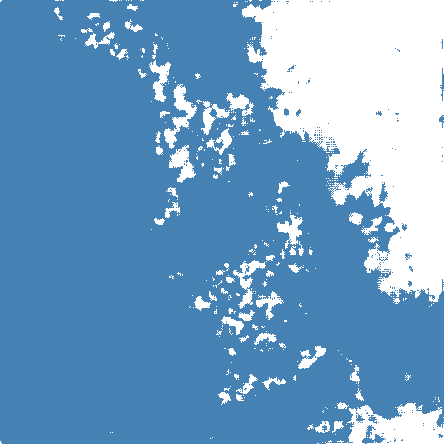} &
    \includegraphics[width=0.12\linewidth]{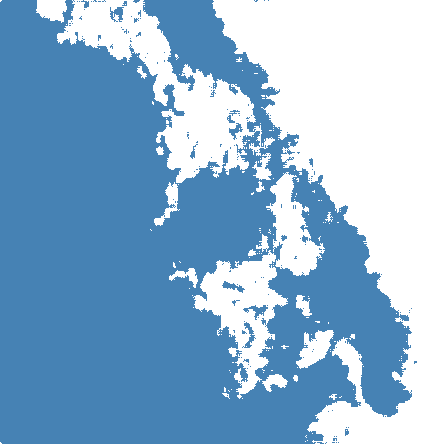} &
    \includegraphics[width=0.12\linewidth]{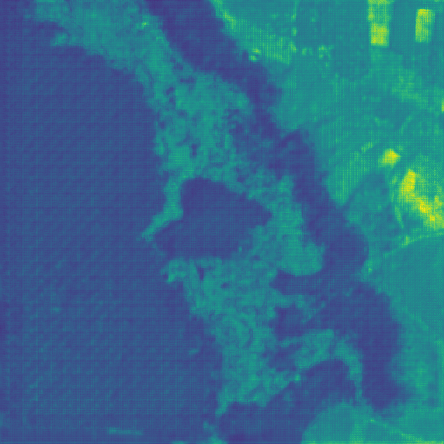} &
    \includegraphics[width=0.12\linewidth]{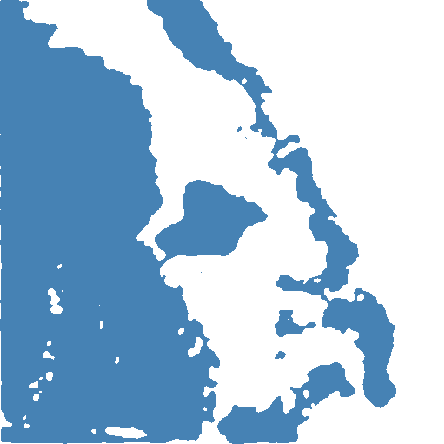}\\
    \includegraphics[width=0.12\linewidth]{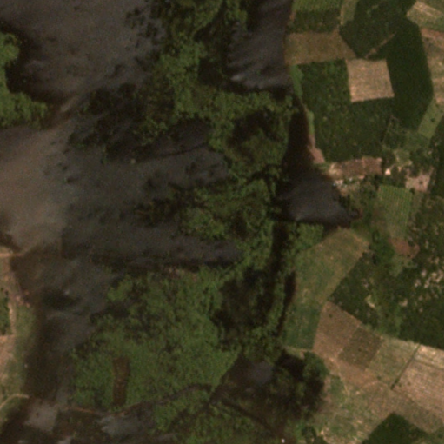} &   \includegraphics[width=0.12\linewidth]{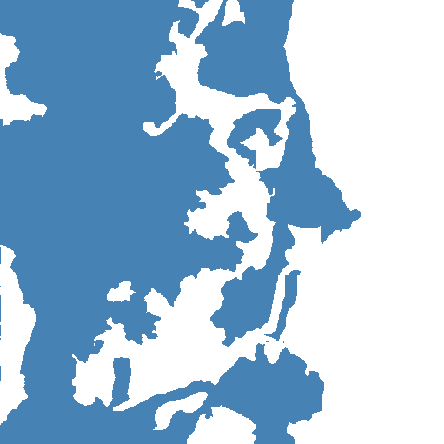} &   \includegraphics[width=0.12\linewidth]{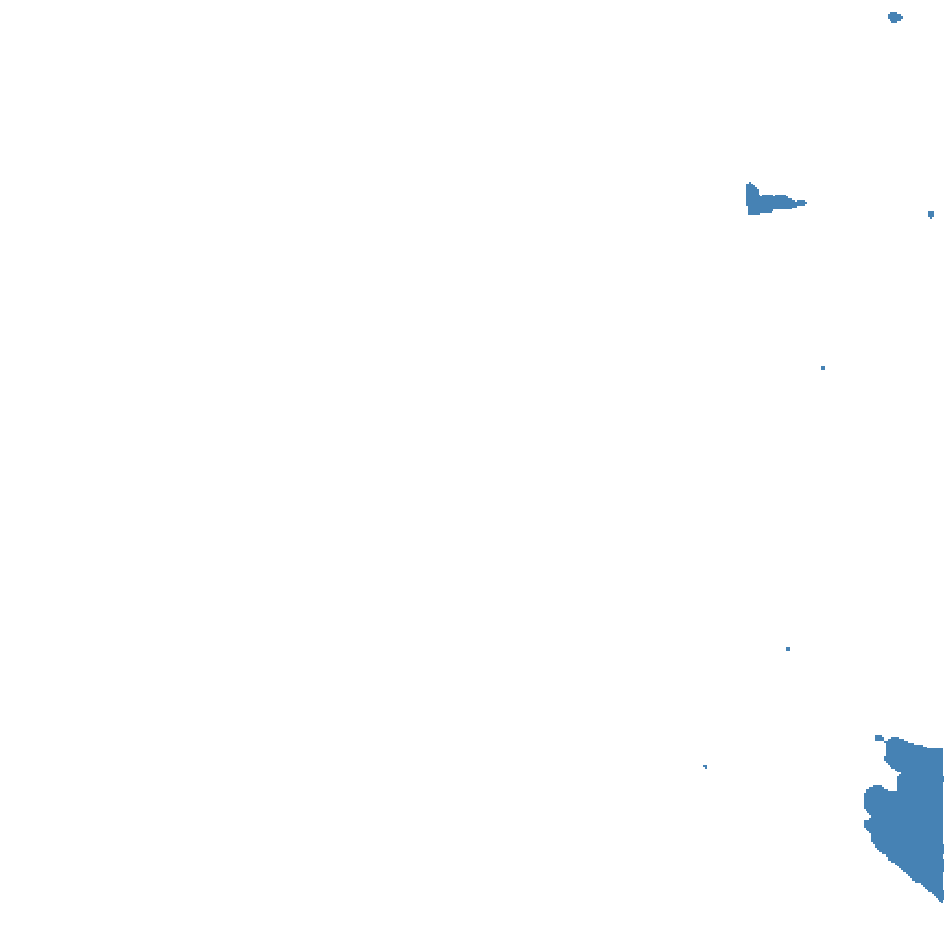} & 
    \includegraphics[width=0.12\linewidth]{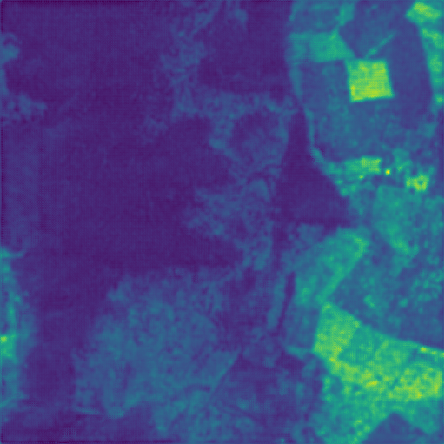} & 
    \includegraphics[width=0.12\linewidth]{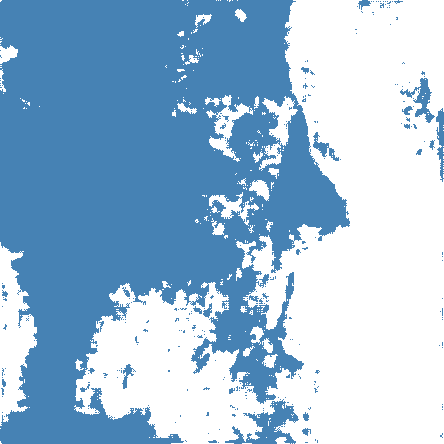} &
    \includegraphics[width=0.12\linewidth]{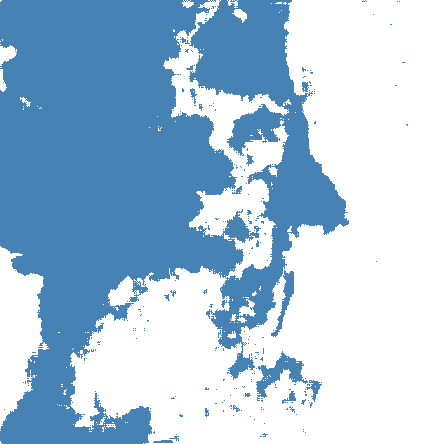} &
    \includegraphics[width=0.12\linewidth]{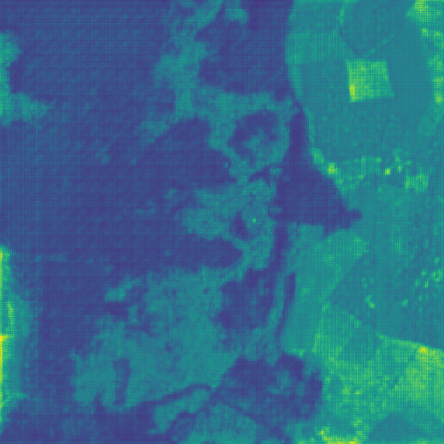} &
    \includegraphics[width=0.12\linewidth]{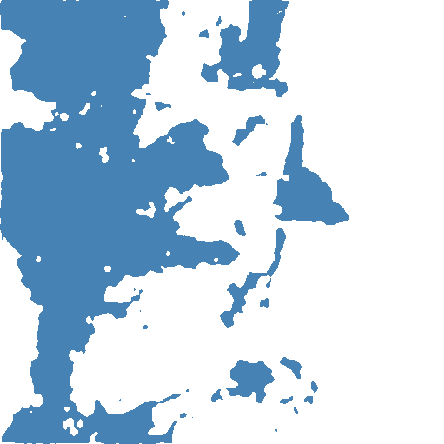}\\
    \includegraphics[width=0.12\linewidth]{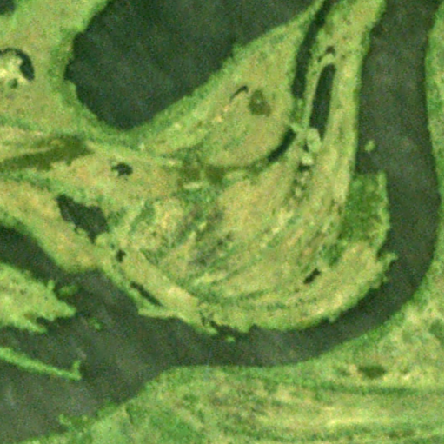} &   \includegraphics[width=0.12\linewidth]{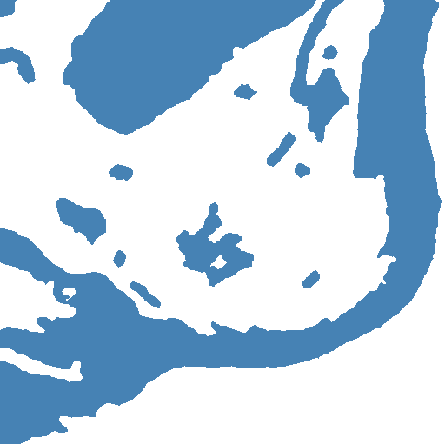} &   \includegraphics[width=0.12\linewidth]{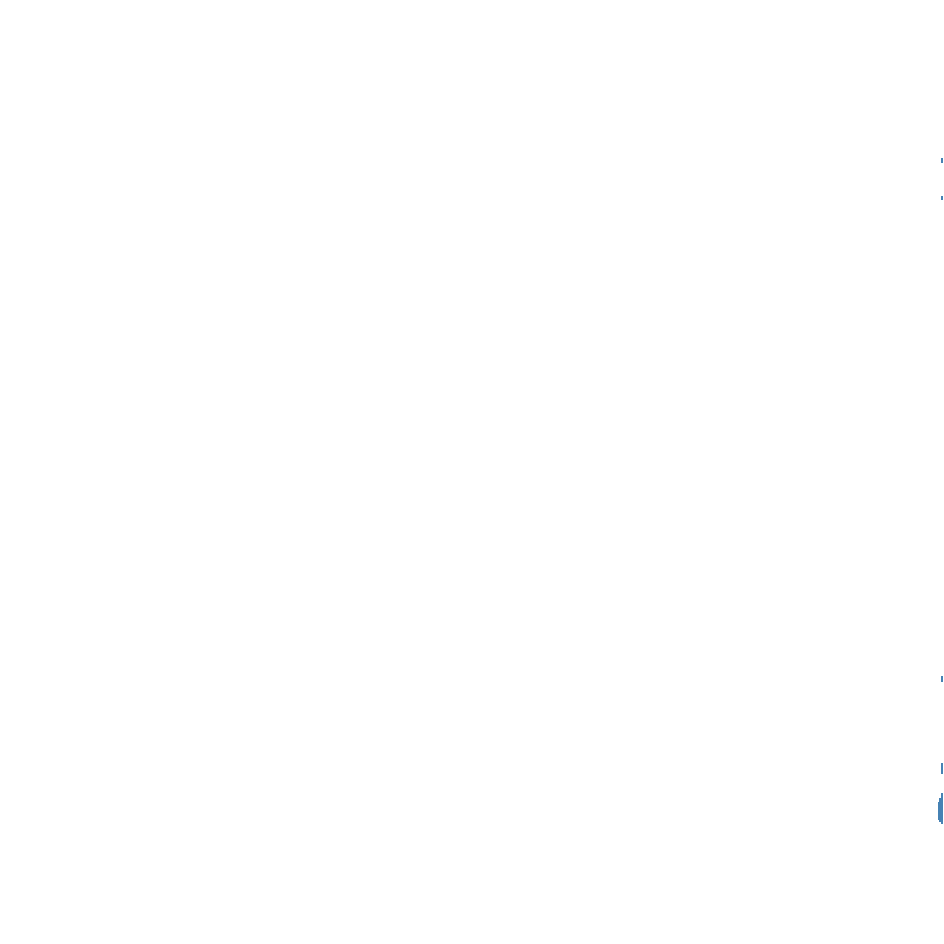} & 
    \includegraphics[width=0.12\linewidth]{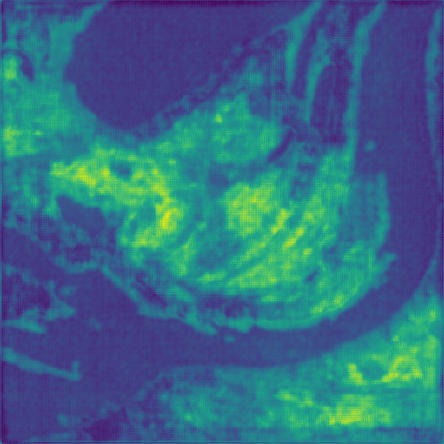} & 
    \includegraphics[width=0.12\linewidth]{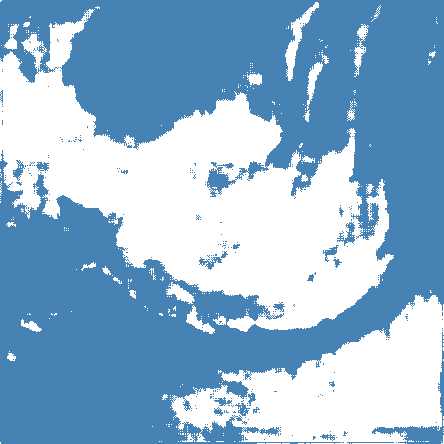} &
    \includegraphics[width=0.12\linewidth]{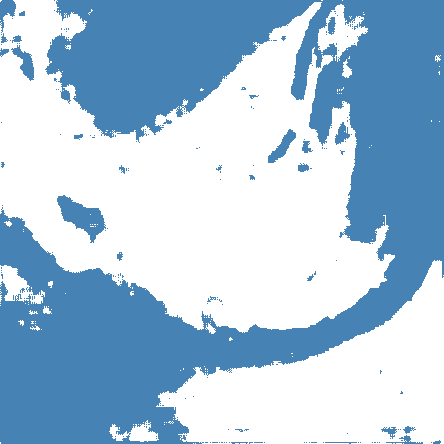} &
    \includegraphics[width=0.12\linewidth]{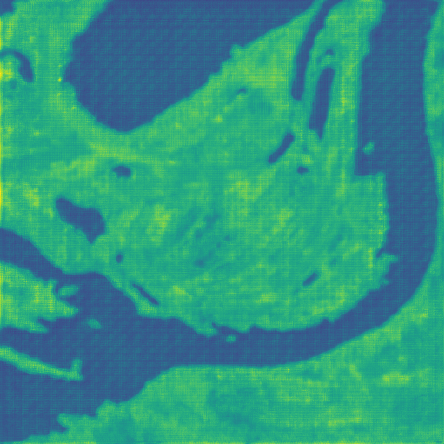} &
    \includegraphics[width=0.12\linewidth]{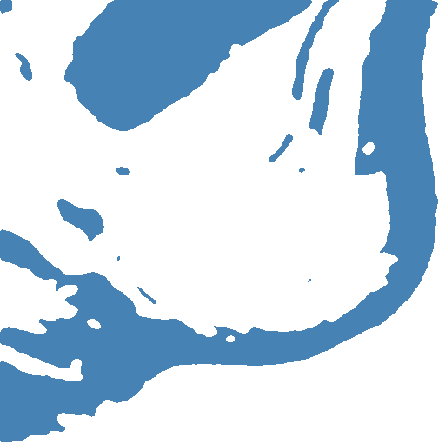}\\
    \includegraphics[width=0.12\linewidth]{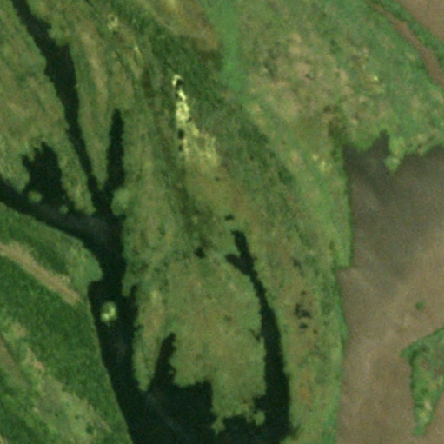} &   \includegraphics[width=0.12\linewidth]{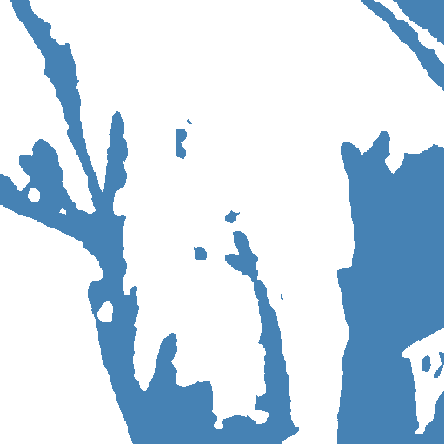} &   \includegraphics[width=0.12\linewidth]{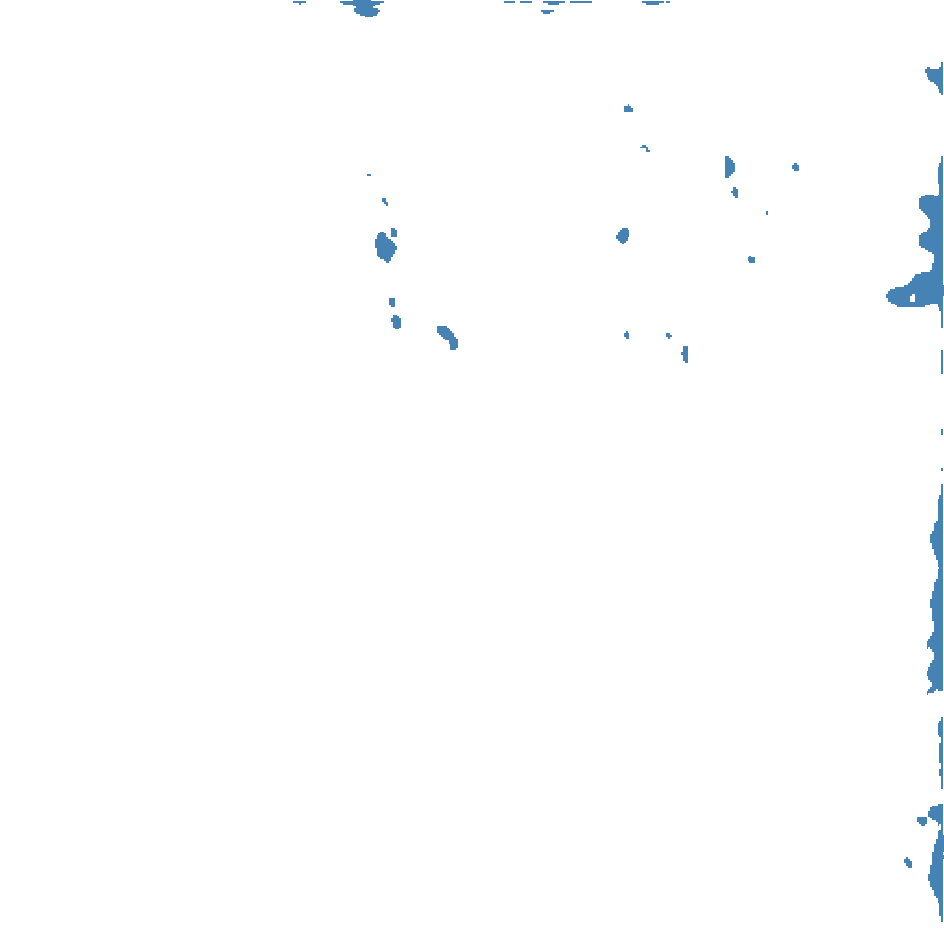} & 
    \includegraphics[width=0.12\linewidth]{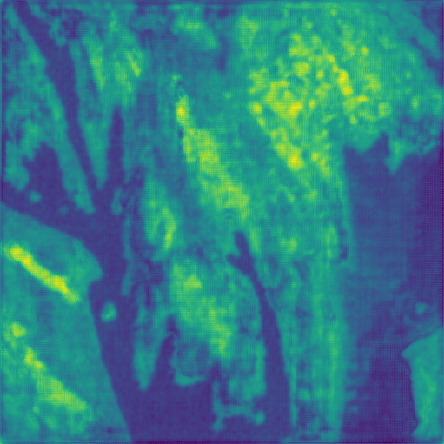} & 
    \includegraphics[width=0.12\linewidth]{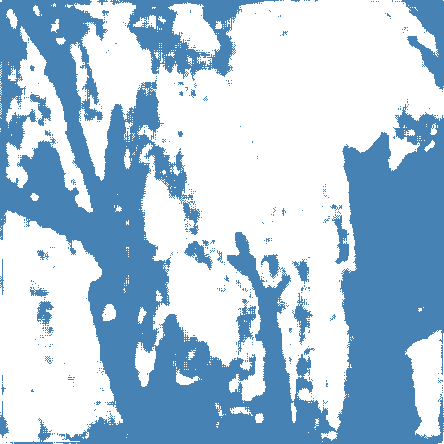} &
    \includegraphics[width=0.12\linewidth]{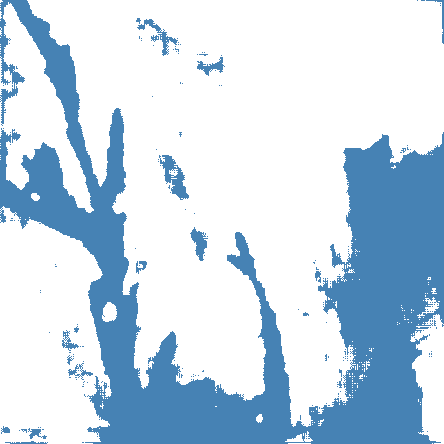} &
    \includegraphics[width=0.12\linewidth]{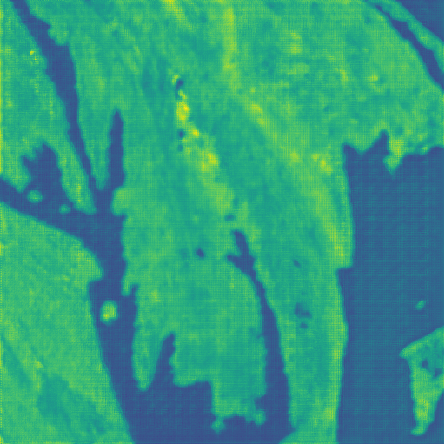} &
    \includegraphics[width=0.12\linewidth]{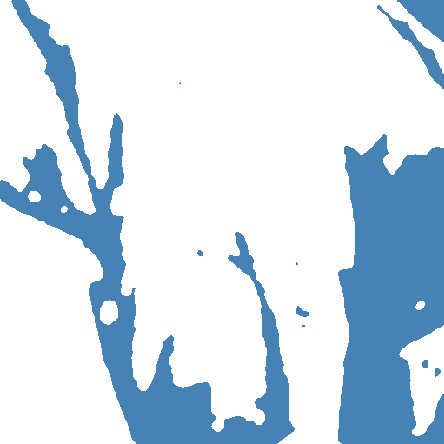}\\

    Input Image & Ground Truth  & \multicolumn{1}{m{2cm}}{\centering U-Net w/ Refiner} & SWIR-Synth & \multicolumn{1}{m{2cm}}{\centering SWIR-Synth Thresh.}  & \multicolumn{1}{m{2cm}}{\centering H2O-Net (RGB)} & \multicolumn{1}{m{2cm}}{\centering SWIR-Synth (RGB+NIR)} & \multicolumn{1}{m{2cm}}{\centering H2O-Net (RGB+NIR)}
\end{tabular}
\vspace{-0.5em}
\caption{Qualitative comparison with SOTA methods on PlanetScope \cite{planet_2020}. H2O-Net results shows that synthesizing SWIR data allows robust segmentation performance in high resolution imagery. It can be seen that adding NIR signals to training and inference improves results in both for segmentation and synthesized SWIR. Blue and white predictions correspond to water and non-water pixels. Best viewed in color and zoomed.}
\vspace{-1.8em}
\label{supp:qual_results}
\end{figure*}

\begin{figure*}[h!]
\setlength\tabcolsep{1.5pt}
\def\arraystretch{1}
\centering
\begin{tabular}{ccccccc}
    \includegraphics[width=0.15\linewidth]{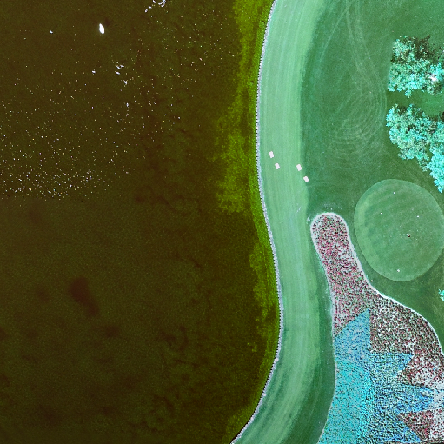} & \includegraphics[width=0.15\linewidth]{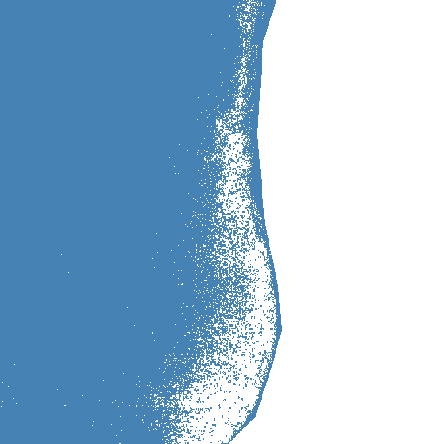} &   \includegraphics[width=0.15\linewidth]{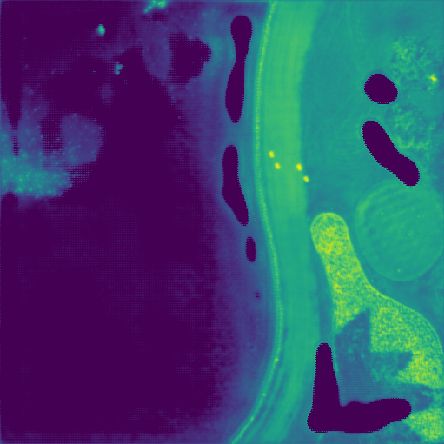} & 
    \includegraphics[width=0.15\linewidth]{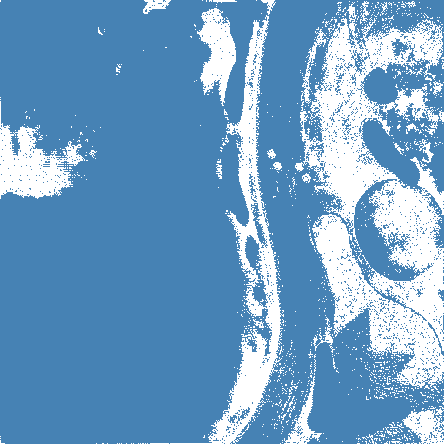} & 
    \includegraphics[width=0.15\linewidth]{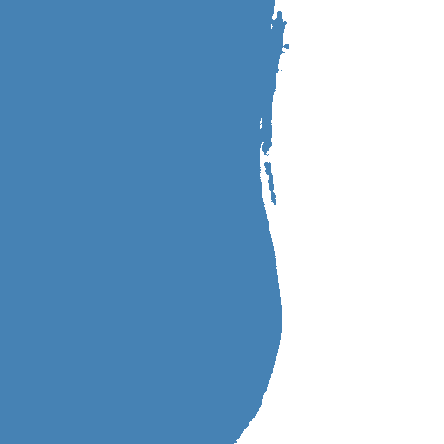} \\
    \includegraphics[width=0.15\linewidth]{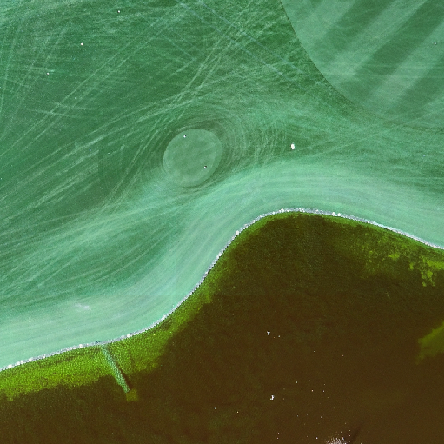} & \includegraphics[width=0.15\linewidth]{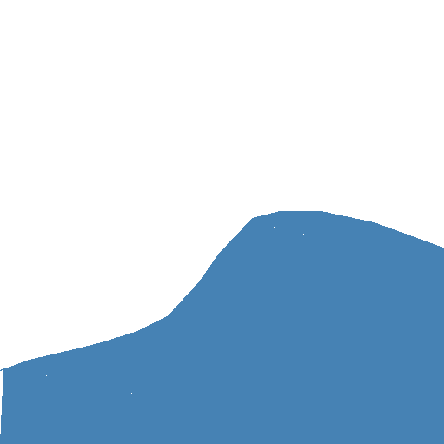} &   \includegraphics[width=0.15\linewidth]{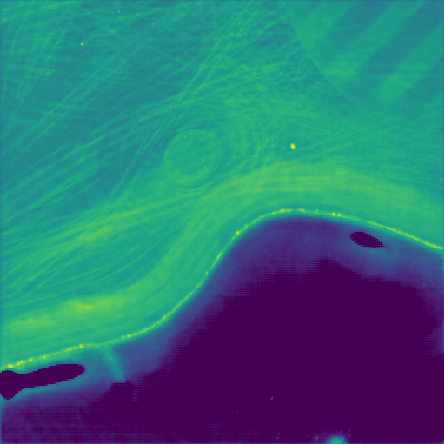} & 
    \includegraphics[width=0.15\linewidth]{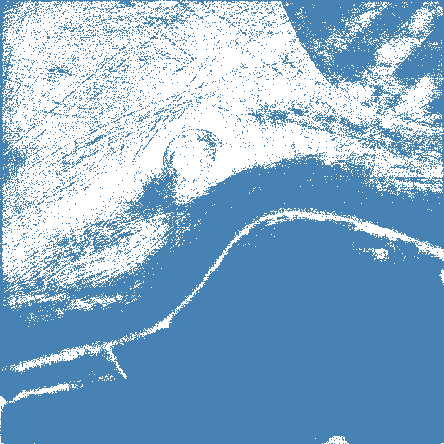} & 
    \includegraphics[width=0.15\linewidth]{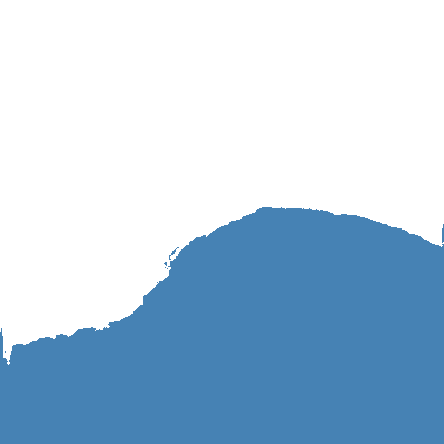} \\
    \includegraphics[width=0.15\linewidth]{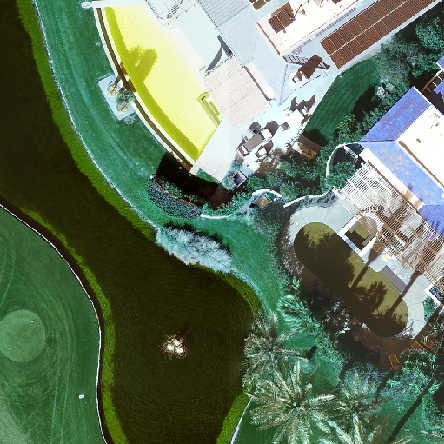} & \includegraphics[width=0.15\linewidth]{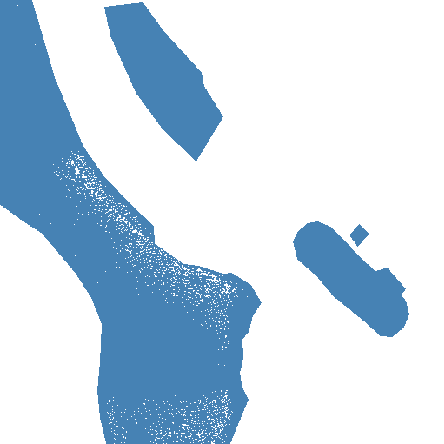} &   \includegraphics[width=0.15\linewidth]{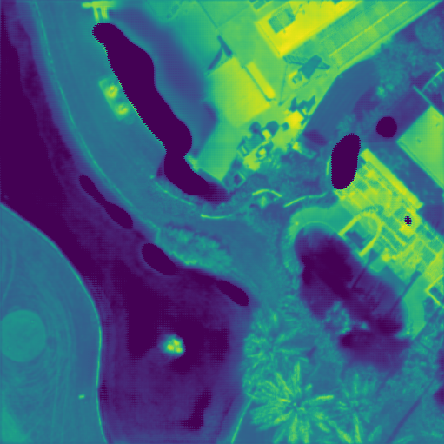} & 
    \includegraphics[width=0.15\linewidth]{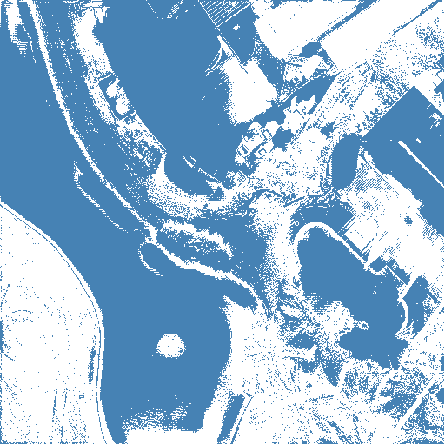} & 
    \includegraphics[width=0.15\linewidth]{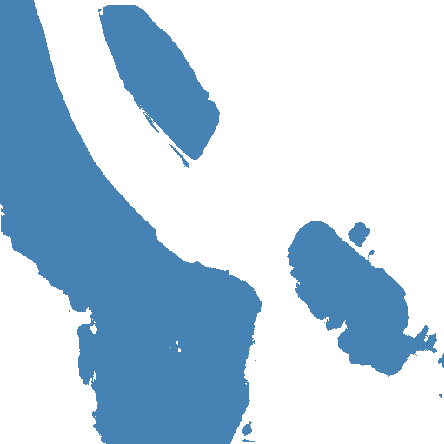} \\
    \includegraphics[width=0.15\linewidth]{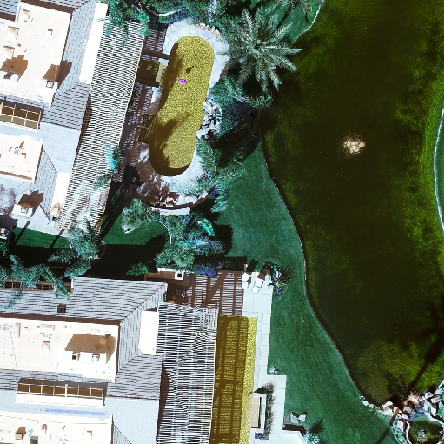} & \includegraphics[width=0.15\linewidth]{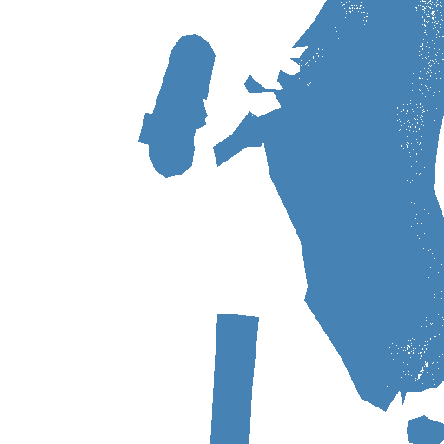} &   \includegraphics[width=0.15\linewidth]{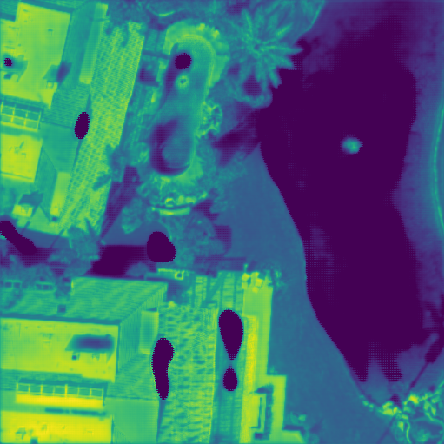} & 
    \includegraphics[width=0.15\linewidth]{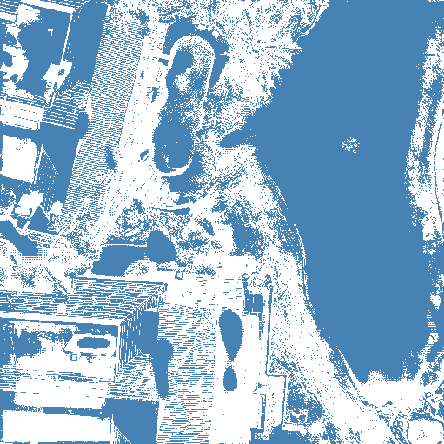} & 
    \includegraphics[width=0.15\linewidth]{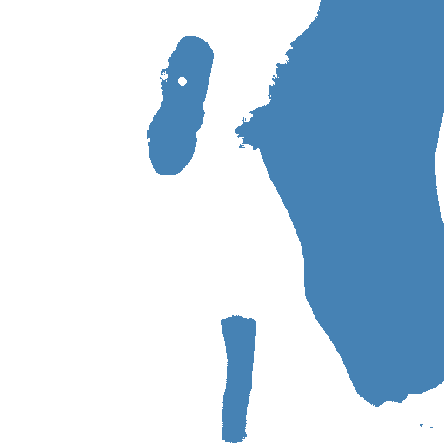} \\
    \includegraphics[width=0.15\linewidth]{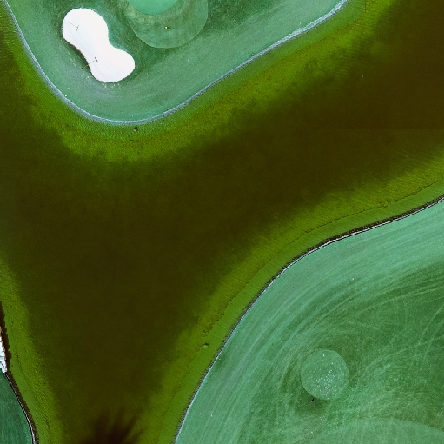} & \includegraphics[width=0.15\linewidth]{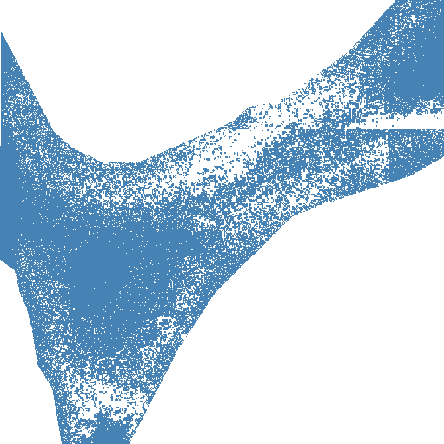} &   \includegraphics[width=0.15\linewidth]{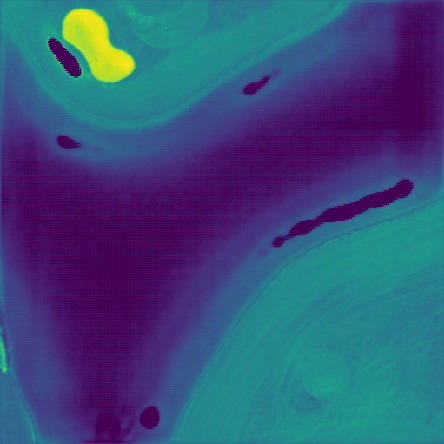} & 
    \includegraphics[width=0.15\linewidth]{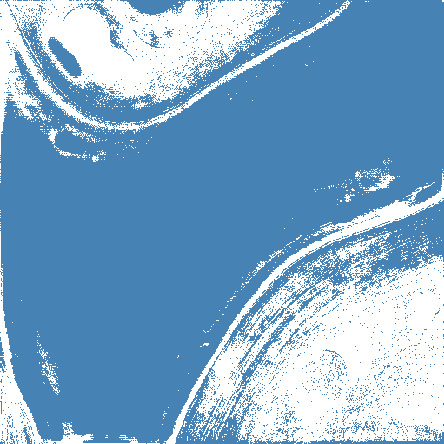} & 
    \includegraphics[width=0.15\linewidth]{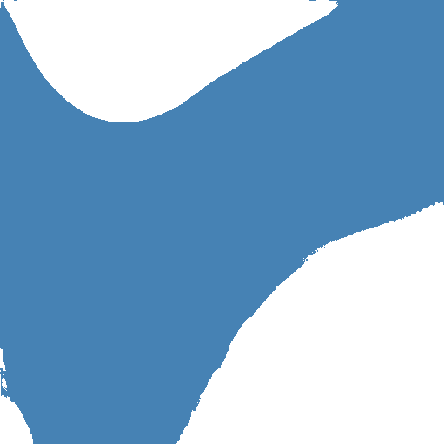} \\
   \includegraphics[width=0.15\linewidth]{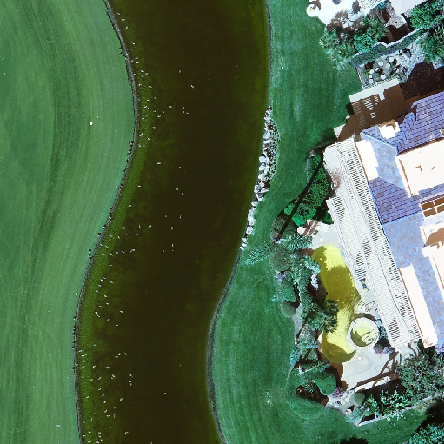} & \includegraphics[width=0.15\linewidth]{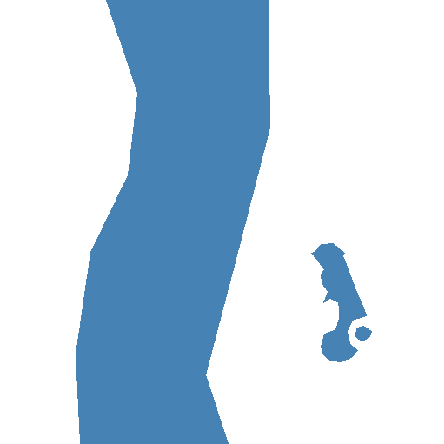} &   \includegraphics[width=0.15\linewidth]{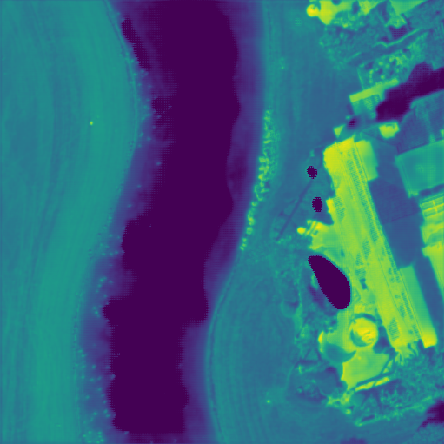} & 
    \includegraphics[width=0.15\linewidth]{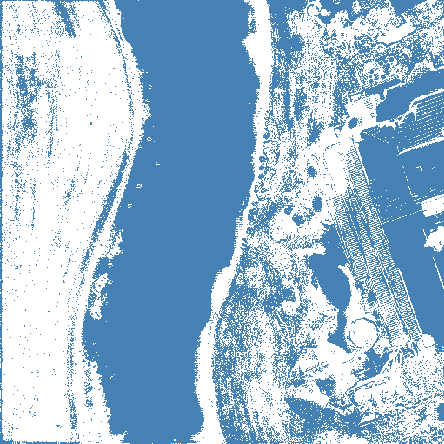} & 
    \includegraphics[width=0.15\linewidth]{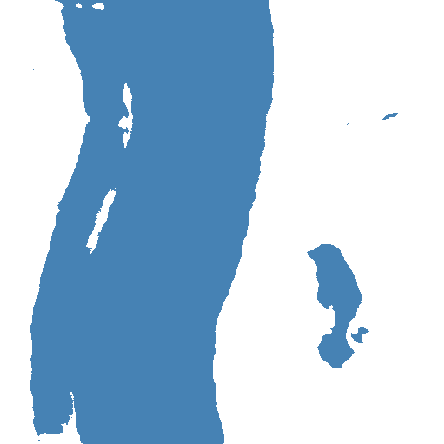} \\
    \includegraphics[width=0.15\linewidth]{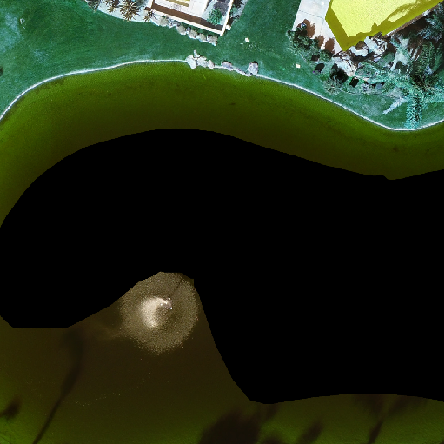} & \includegraphics[width=0.15\linewidth]{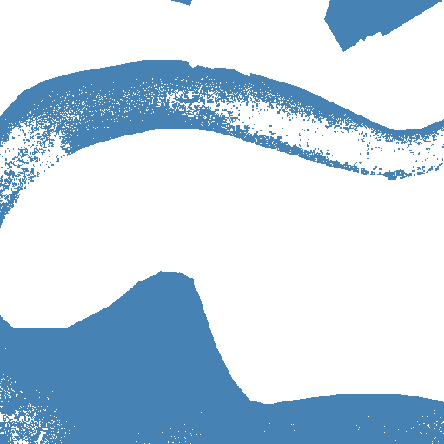} &   \includegraphics[width=0.15\linewidth]{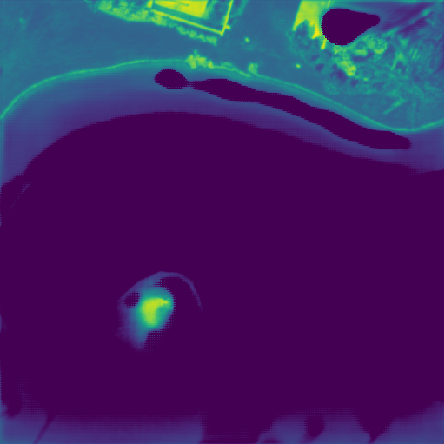} & 
    \includegraphics[width=0.15\linewidth]{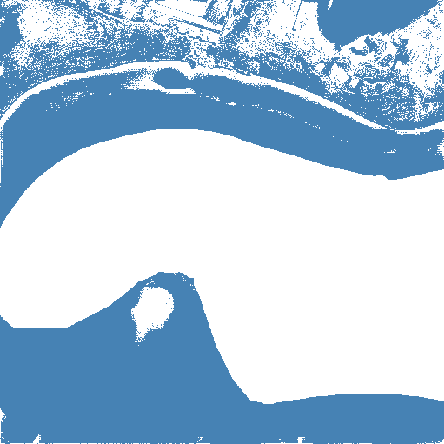} & 
    \includegraphics[width=0.15\linewidth]{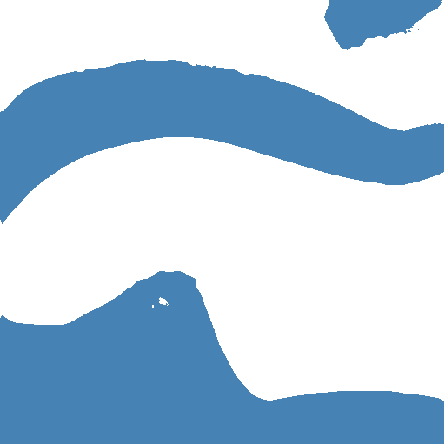} \\
    Input Image & Ground Truth & SWIR-Synth &  SWIR-Synth Threshold & H2O-Net (Ours) 
\end{tabular}
\vspace{-0.2cm}
\caption{Qualitative comparison with SOTA methods on Drone Deploy. Our method (H2O-Net) shows that synthesizing SWIR data allows robust segmentation performance in high resolution imagery. Best viewed in color and zoomed.}
\vspace{-0.4cm}
\label{supp:dd_qual}
\end{figure*}

\begin{figure*}[h!]
\setlength\tabcolsep{1.5pt}
\def\arraystretch{1}
\centering
\begin{tabular}{cccc}
    \includegraphics[width=0.2\linewidth]{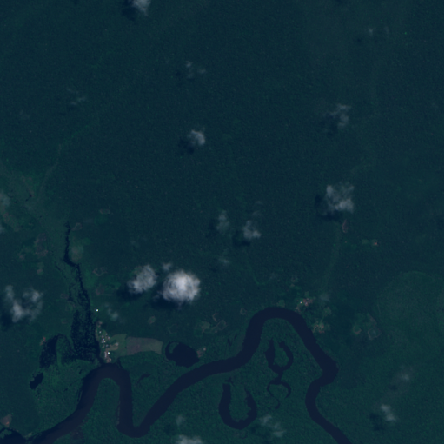} &   \includegraphics[width=0.2\linewidth]{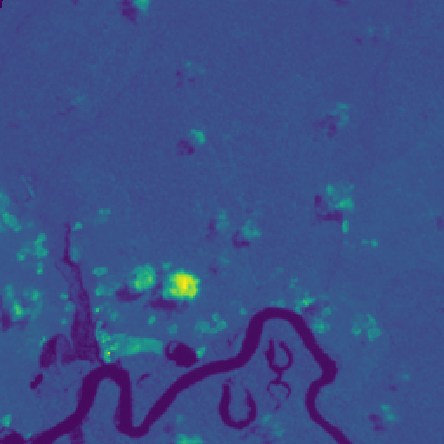} &   
    \includegraphics[width=0.2\linewidth]{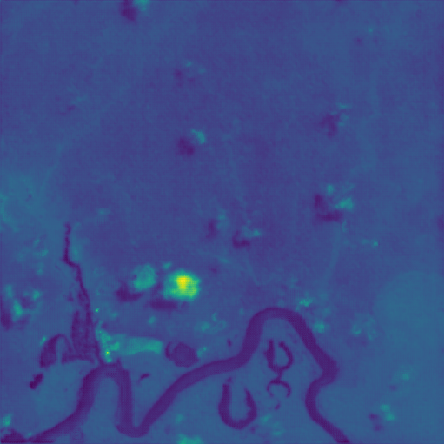} \\
    \includegraphics[width=0.2\linewidth]{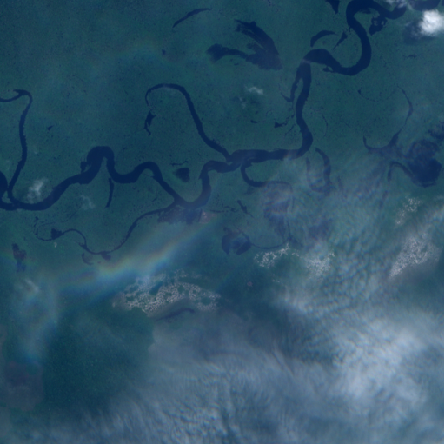} & 
    \includegraphics[width=0.2\linewidth]{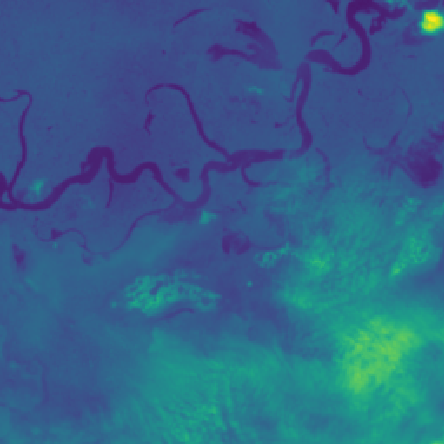} &    
    \includegraphics[width=0.2\linewidth]{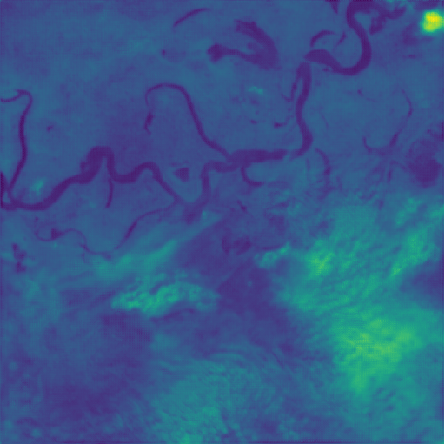} \\
    \includegraphics[width=0.2\linewidth]{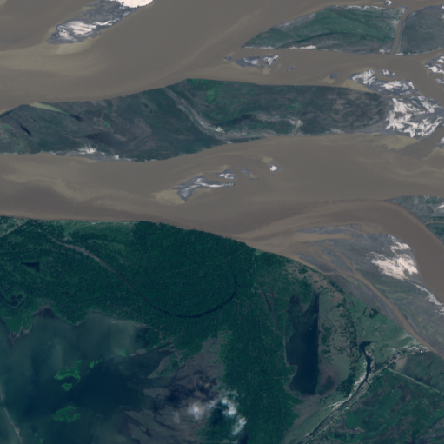} & 
    \includegraphics[width=0.2\linewidth]{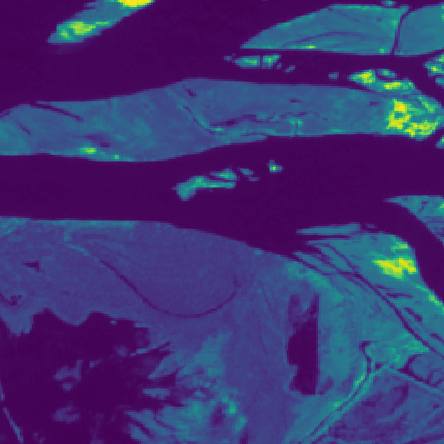} &    
    \includegraphics[width=0.2\linewidth]{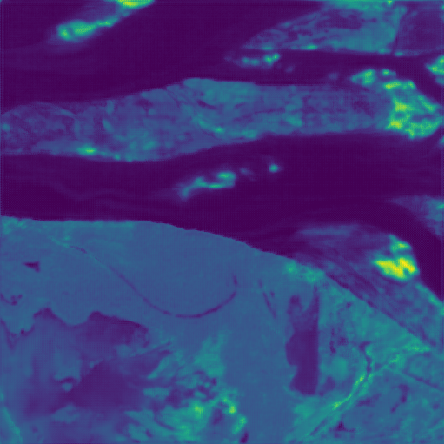}\\
    \includegraphics[width=0.2\linewidth]{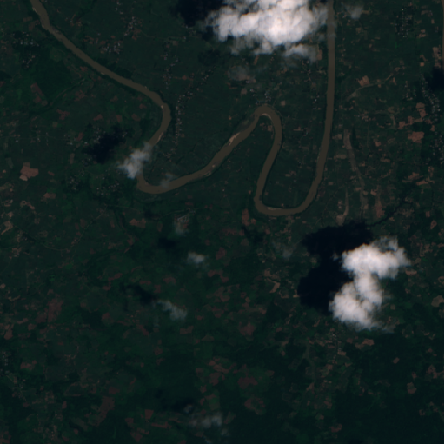} & 
    \includegraphics[width=0.2\linewidth]{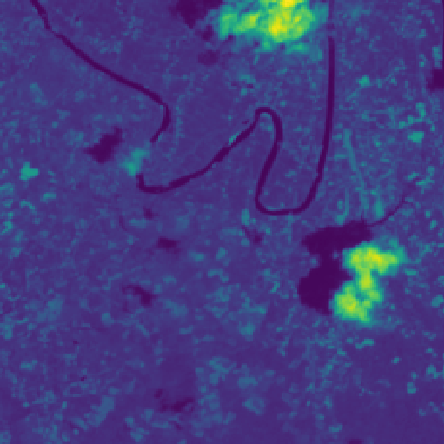} &    
    \includegraphics[width=0.2\linewidth]{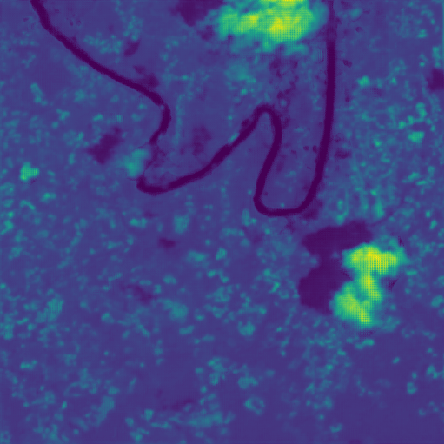} \\
    \includegraphics[width=0.2\linewidth]{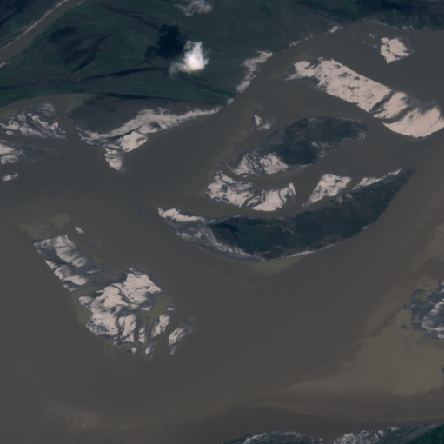} & 
    \includegraphics[width=0.2\linewidth]{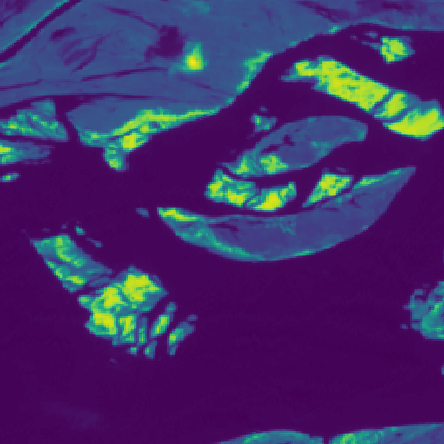} &    
    \includegraphics[width=0.2\linewidth]{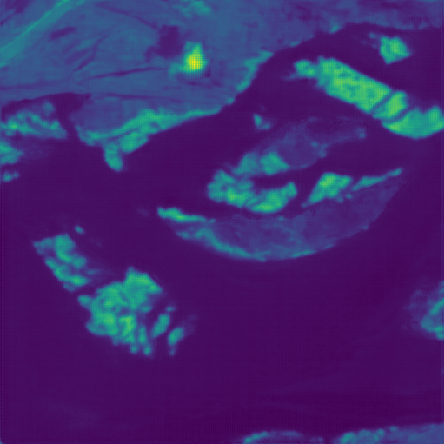} \\
    \includegraphics[width=0.2\linewidth]{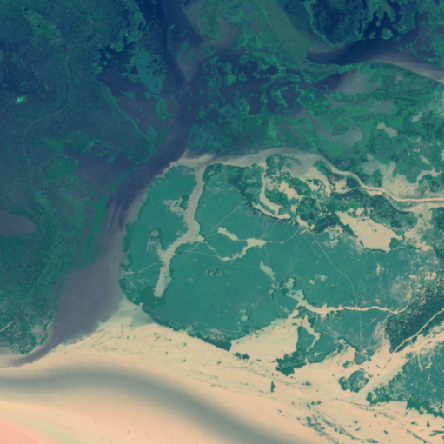} & 
    \includegraphics[width=0.2\linewidth]{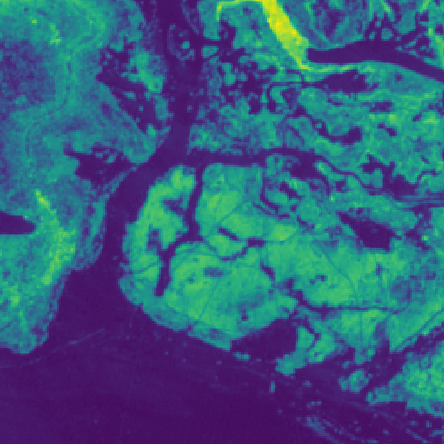} &   
    \includegraphics[width=0.2\linewidth]{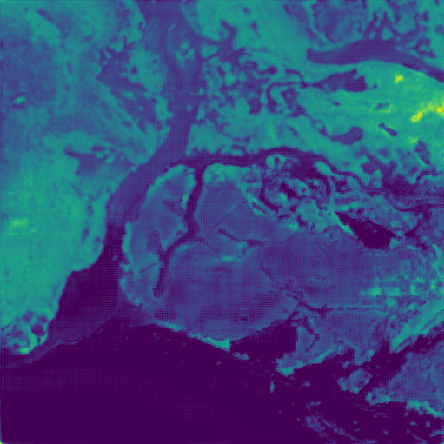} \\
    Input Image & Real SWIR_\textsubscript{2} & SWIR-Synth 
\end{tabular}
\vspace{-0.2cm}
\caption{Qualitative results of SWIR-synth and ground truth SWIR\textsubscript{2} in low resolution imagery, Sentinel-2. Best viewed in color and zoomed.}
\vspace{-0.4cm}
\label{fig:low_res_swir_synth}
\end{figure*}

\begin{figure*}[h!]
\setlength\tabcolsep{1.5pt}
\def\arraystretch{1}
\centering
\begin{tabular}{cccc}
    \includegraphics[width=0.2\linewidth]{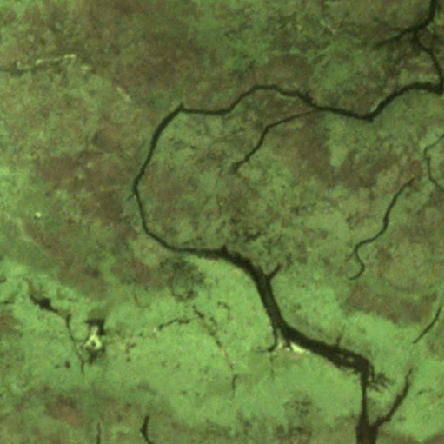} &   \includegraphics[width=0.2\linewidth]{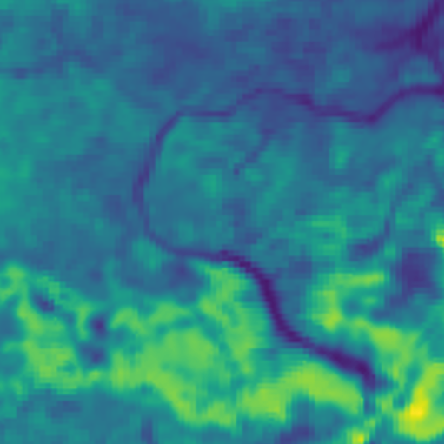} &   
    \includegraphics[width=0.2\linewidth]{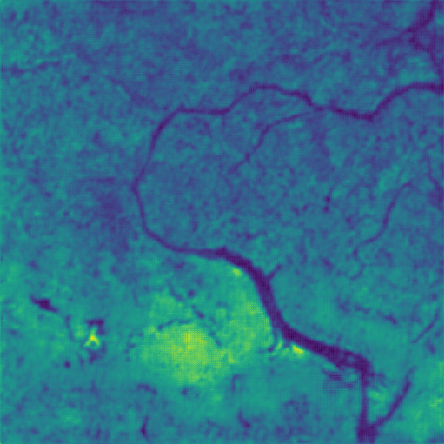} &
    \includegraphics[width=0.2\linewidth]{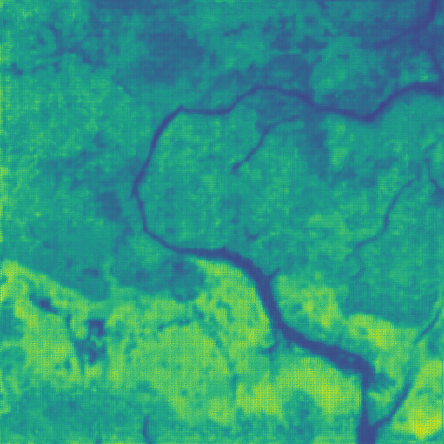} \\
    \includegraphics[width=0.2\linewidth]{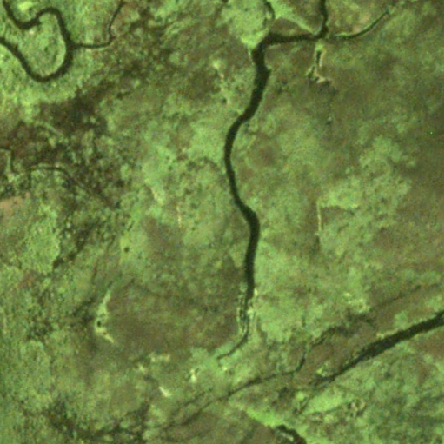} & \includegraphics[width=0.2\linewidth]{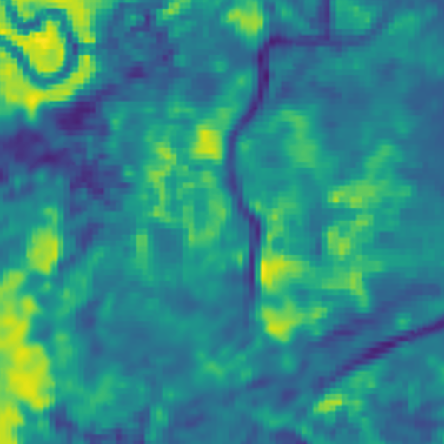} &    
    \includegraphics[width=0.2\linewidth]{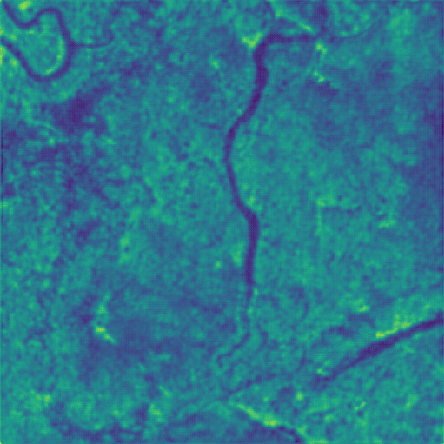} &
    \includegraphics[width=0.2\linewidth]{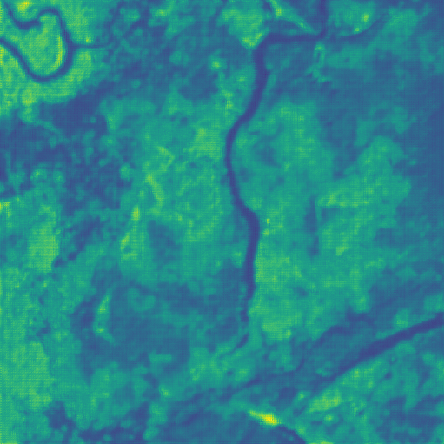} \\
    \includegraphics[width=0.2\linewidth]{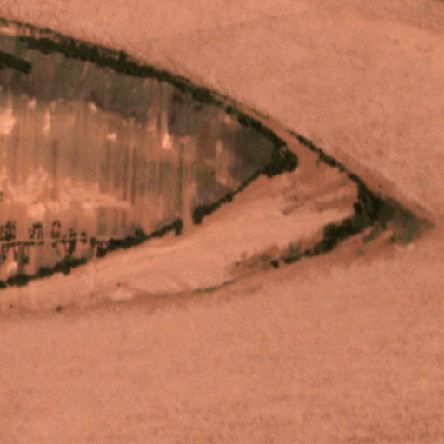} & \includegraphics[width=0.2\linewidth]{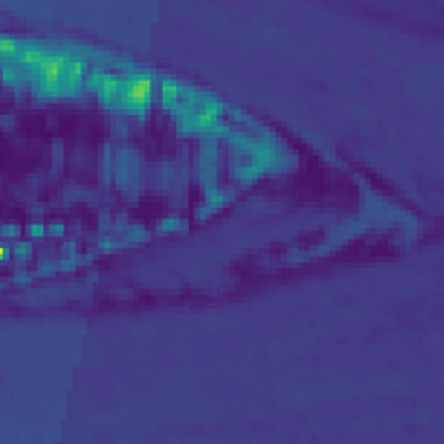} &    
    \includegraphics[width=0.2\linewidth]{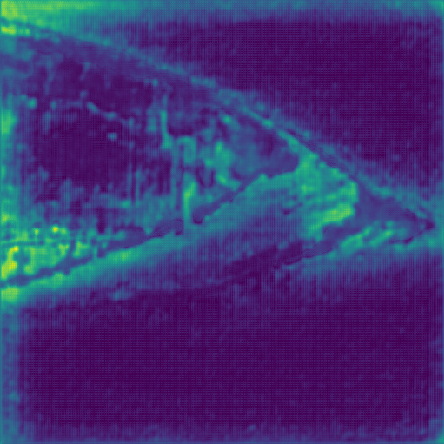} &
    \includegraphics[width=0.2\linewidth]{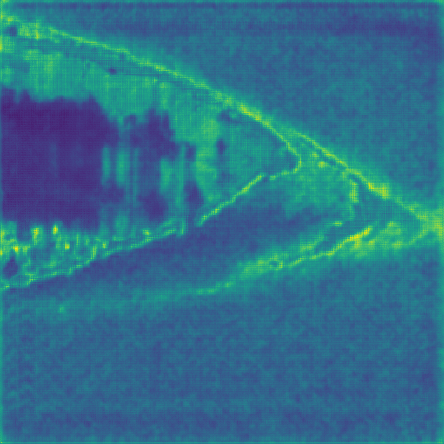} \\
    \includegraphics[width=0.2\linewidth]{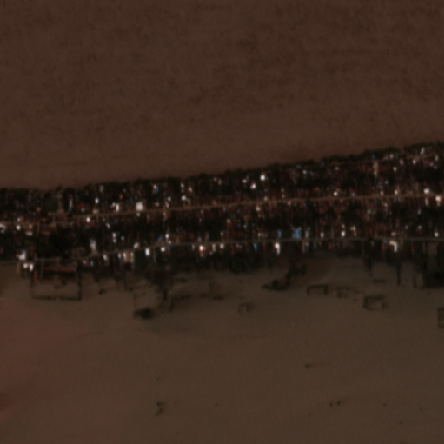} & \includegraphics[width=0.2\linewidth]{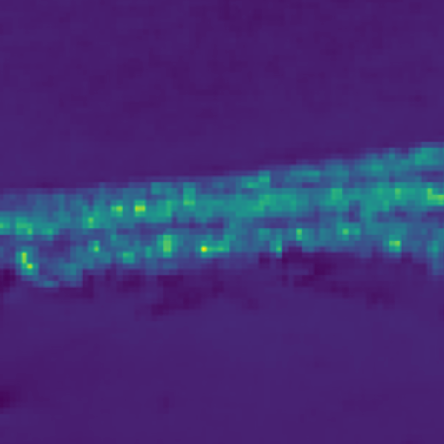} &    
    \includegraphics[width=0.2\linewidth]{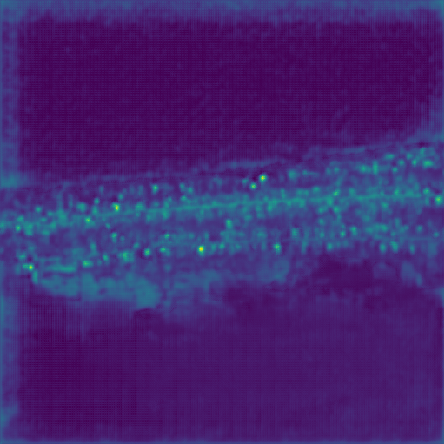} &
    \includegraphics[width=0.2\linewidth]{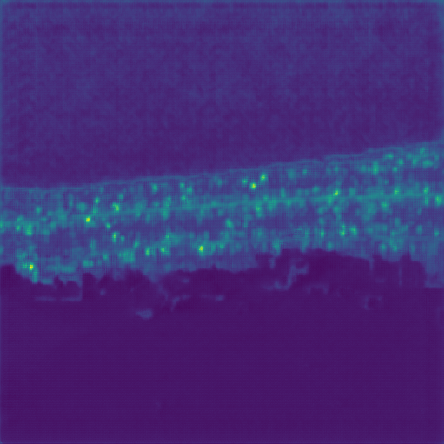} \\
    \includegraphics[width=0.2\linewidth]{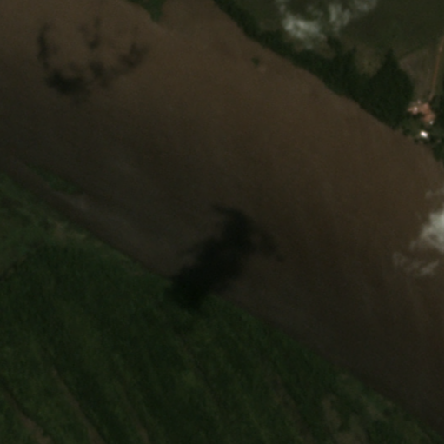} & \includegraphics[width=0.2\linewidth]{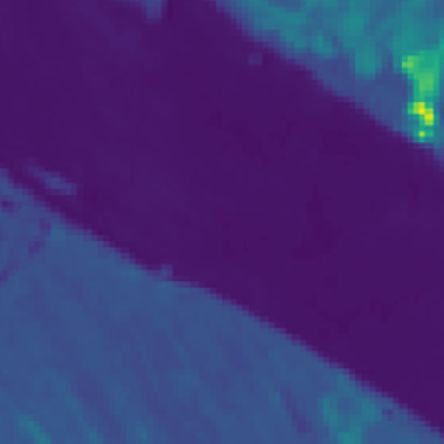} &   
    \includegraphics[width=0.2\linewidth]{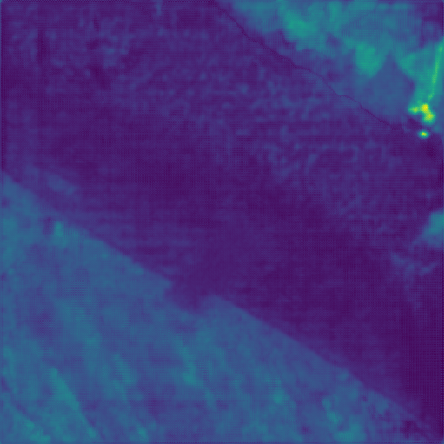} &
    \includegraphics[width=0.2\linewidth]{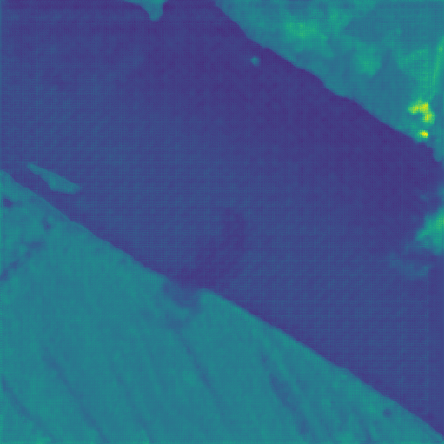} \\
    \includegraphics[width=0.2\linewidth]{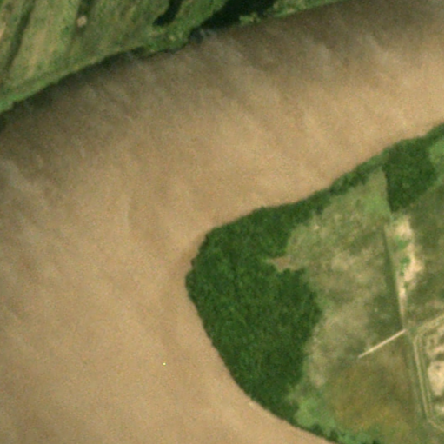} & \includegraphics[width=0.2\linewidth]{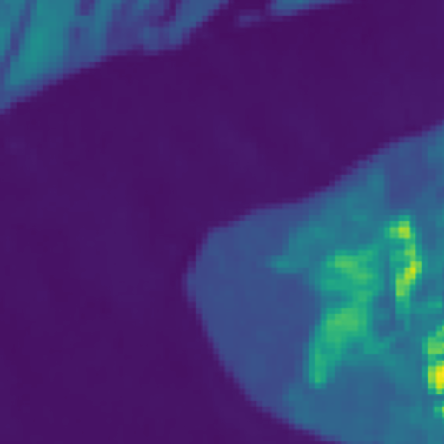} &  
    \includegraphics[width=0.2\linewidth]{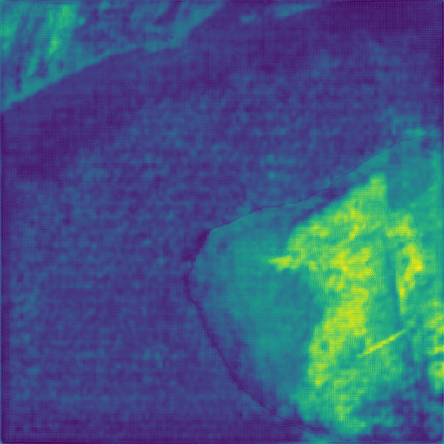} &
    \includegraphics[width=0.2\linewidth]{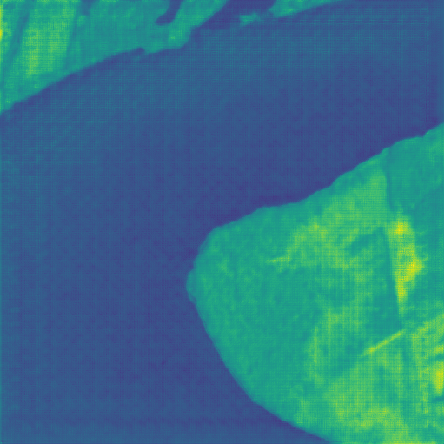} \\
    Input Image & *Real SWIR_$\textsubscript{2}$ & SWIR-Synth (RGB) & SWIR-Synth (RGB+NIR) 
\end{tabular}
\vspace{-0.2cm}
\caption{Qualitative results of SWIR-synth and *real SWIR\textsubscript{2} in high resolution imagery, PlanetScope. *Note that real SWIR\textsubscript{2} here is obtained through scarce overlapping crops with low resolution data. Further details is described in section \ref{section:datasets} in the main paper. Best viewed in color and zoomed.}
\vspace{-0.4cm}
\label{fig:high_res_swir_synth}
\end{figure*}

\begin{figure*}[h!]
\setlength\tabcolsep{1.5pt}
\def\arraystretch{1}
\centering
\begin{tabular}{ccccccc}
    \includegraphics[width=0.16\linewidth]{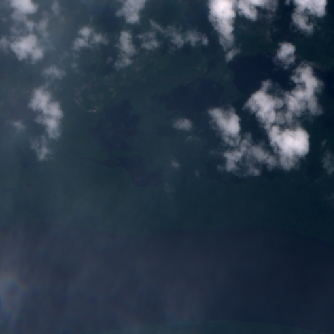} &   \includegraphics[width=0.16\linewidth]{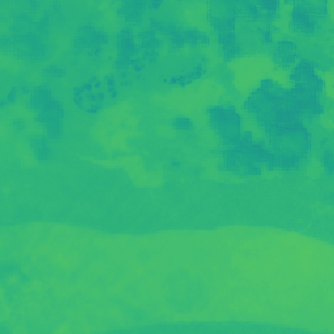} &   \includegraphics[width=0.16\linewidth]{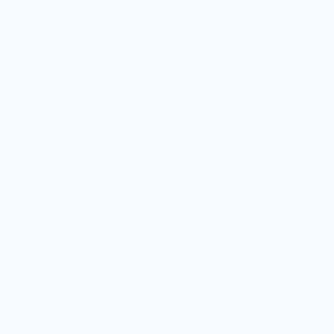} & 
    \includegraphics[width=0.16\linewidth]{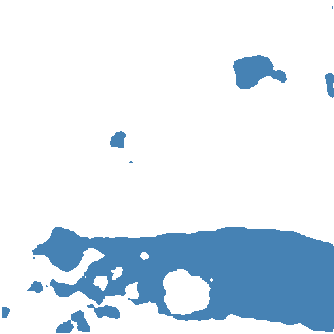} & 
    \includegraphics[width=0.16\linewidth]{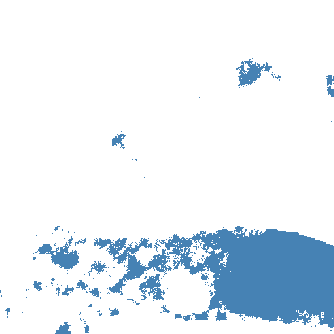} \\
    \includegraphics[width=0.16\linewidth]{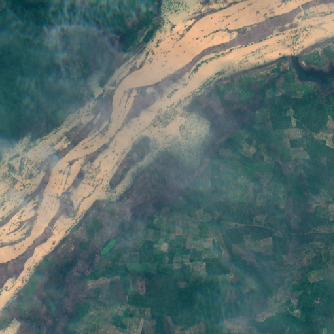} & \includegraphics[width=0.16\linewidth]{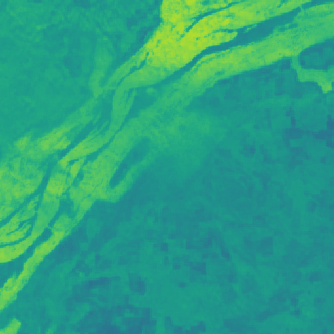} &   \includegraphics[width=0.16\linewidth]{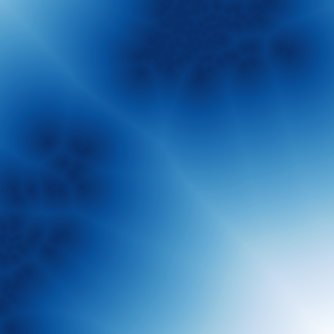} &  
    \includegraphics[width=0.16\linewidth]{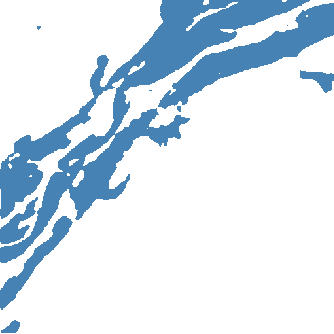} & 
    \includegraphics[width=0.16\linewidth]{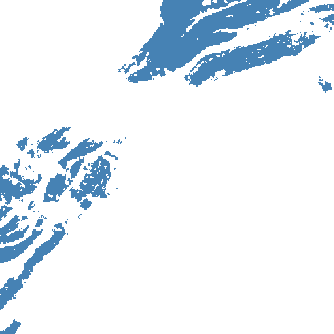} \\
    \includegraphics[width=0.16\linewidth]{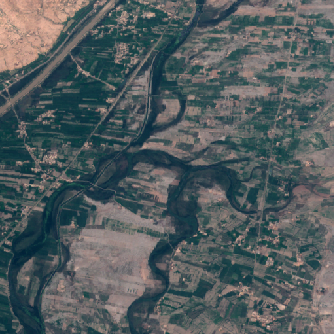} & \includegraphics[width=0.16\linewidth]{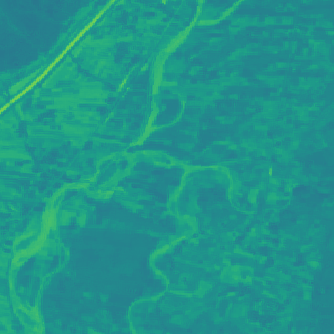} &   \includegraphics[width=0.16\linewidth]{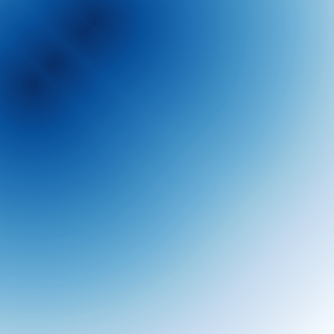} &  
    \includegraphics[width=0.16\linewidth]{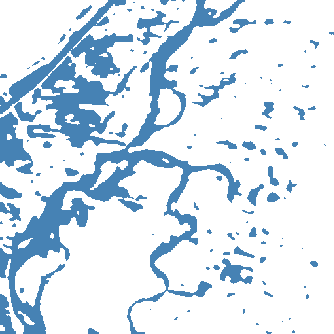} &
    \includegraphics[width=0.16\linewidth]{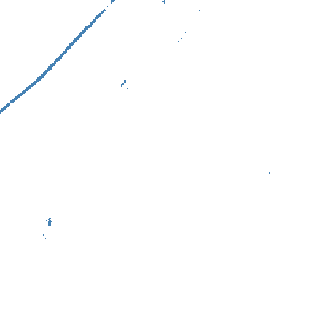} \\
    \includegraphics[width=0.16\linewidth]{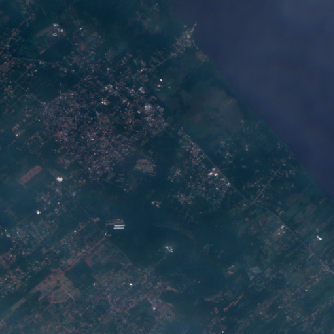} & \includegraphics[width=0.16\linewidth]{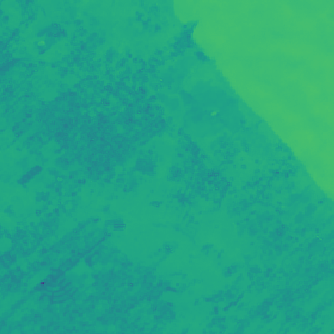} &   \includegraphics[width=0.16\linewidth]{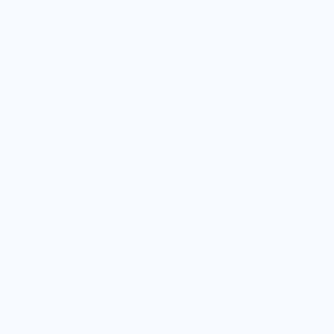} &  
    \includegraphics[width=0.16\linewidth]{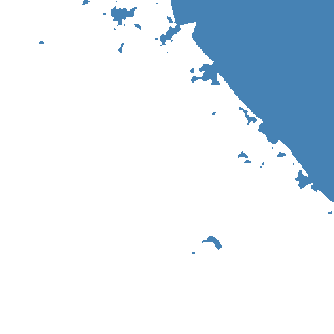} & 
    \includegraphics[width=0.16\linewidth]{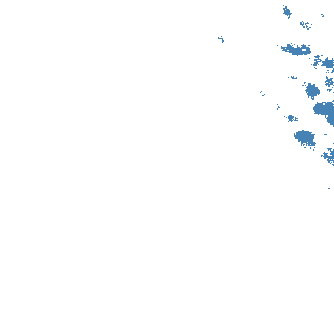} \\
    \includegraphics[width=0.16\linewidth]{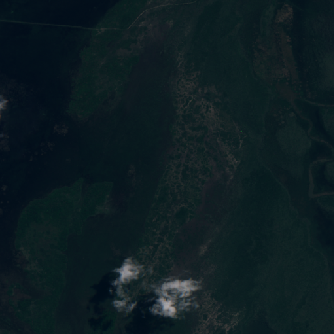} & \includegraphics[width=0.16\linewidth]{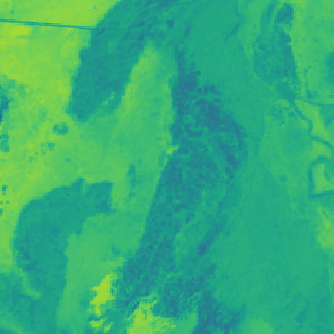} &   \includegraphics[width=0.16\linewidth]{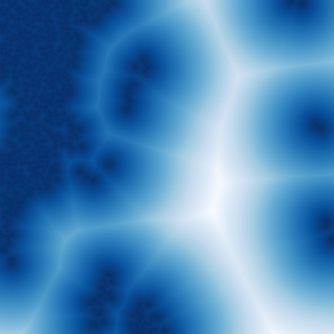} &  
    \includegraphics[width=0.16\linewidth]{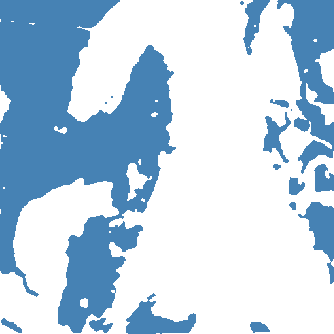} &
    \includegraphics[width=0.16\linewidth]{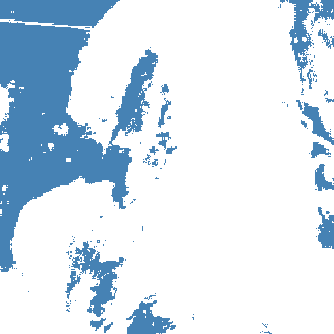} \\
    \includegraphics[width=0.16\linewidth]{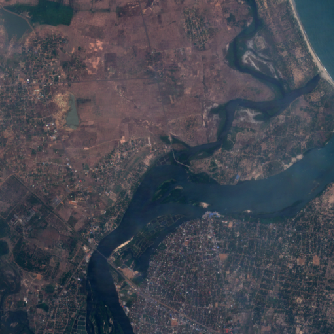} & \includegraphics[width=0.16\linewidth]{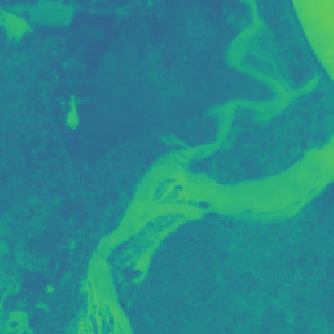} &   \includegraphics[width=0.16\linewidth]{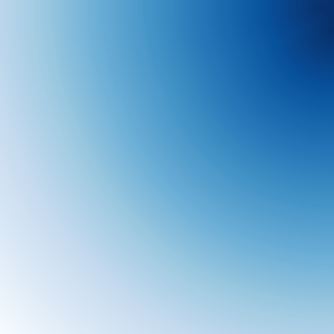} &  
    \includegraphics[width=0.16\linewidth]{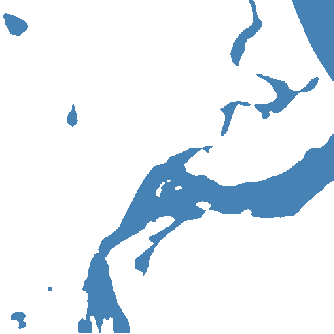} & 
    \includegraphics[width=0.16\linewidth]{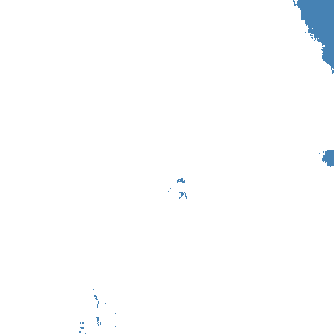} \\
    Input Image & MNDWI & Distance Map & Refined Mask & MNDWI Threshold 
\end{tabular}
\vspace{-0.2cm}
\caption{Qualitative results of the refiner on low resolution data from Sentinel 2 satellite. It can be seen that the refiner often completes hidden parts and refines overall mask. Best viewed in color and zoomed.}
\vspace{-0.4cm}
\label{fig:low_res_refiner}
\end{figure*}

Here we provide additional qualitative result to illustrate performance in each stage of H2O-Net. Figure \ref{fig:low_res_swir_synth} illustrates features SWIR-Synth learns in low resolution data, showing the input image, ground truth SWIR\textsubscript{2}, synthesized SWIR\textsubscript{2}, thresholded mask obtained from ground truth SWIR\textsubscript{2}, and predicted thresholded mask obtained from synthesized SWIR\textsubscript{2}. Figure \ref{fig:high_res_swir_synth} demonstrate SWIR-Synth qualitative performance on high resolution data. Figure \ref{fig:low_res_refiner} show refiner performance on low resolution data. Figure \ref{supp:qual_results} shows additional results on PlanetScope \cite{planet_2020}, and figure \ref{supp:dd_qual} shows qualitative results on DroneDeploy \cite{dronedeploy}.
\section{Implementation Details}
\subsection{SWIR-Synth Network Architecture}
SWIR-SynthNet is structured as an image-to-image translation network. The generator is constructed of 3 encoding blocks, and 3 decoding blocks with skip connections. Each encoding block is comprised of a spectrally normalized convolution layer with a 3$\times$3 kernel and 1 padding, batch normalization layer, leaky rectified linear unit (ReLU) with slope of 0.2 and a dropout layer. Each block output is then down-sampled by a factor of 2 using a strided convolution layer. Decoding blocks first upsample input using a spectrally normalized transposed convolution layer with a 4$\times$4 kernel size and a stride of 2, followed by a spectrally normalized convolution layer with a 3$\times$3 kernel and 1 padding, batch normalization layer, leaky ReLU with slope of 0.2 and a dropout layer. Decoding blocks take a concatenation of the previous decoding block and output of the corresponding encoder block. The discriminator has 5 blocks each with a spectrally normalized convolution layer with 4$\times$4 kernel size, and stride of 2, followed by a batch normalization, and a ReLU activation function. A self-attention module similar to \cite{zhang2019self} was added after the second and forth blocks. The last layer of the last block outputs the features tensor for the feature loss, which is also used as an input to a Sigmoid activation layer for the adversarial loss. 
\subsection{Segmentation Architecture}
 The segmentation network uses an encoder-decoder architecture similar to U-Net \cite{ronneberger2015u}, with 5 encoding blocks and 5 decoding blocks with skip connections. Encoding blocks comprise of two spectrally normalized convolution layers, batch normalization layer, and LeakyReLU layer. Each block is followed by a strided convolution layer acting as a pooling layer. The decoder also has 5 blocks, each comprised of transposed convolution layer for upsampling, two spectrally normalized convolution layers, batch normalization layer, and a Leaky ReLU layer. Skip connections are used for corresponding encoding and decoding blocks. This architecture was also used for the U-Net baseline method to ensure consistency and fairness in evaluation.
\subsection{Refiner Architecture}
has one encoder block and one decoder block. The encoder block consists of 2 convolution layers each followed by a normalization layer, batch normalization layer after first convolution and instance normalization layer after second convolution, and a Leaky ReLU layer. The decoder block first upsamples features using bilinear interpolation, followed by 2 convolution layers, normalization layers, and Leaky ReLU layers. For high confidence points sampling, we use $\phi_H$ of 0.5, and $\phi_L$ of -0.2.


\clearpage
{\small
\bibliographystyle{ieee_fullname}
\bibliography{egbib.bib}
}